\documentclass[preprint,12pt]{elsarticle}

\usepackage{graphicx} % Required for inserting images
\usepackage{amssymb}
\usepackage{amsmath}
\usepackage{url}
\usepackage{booktabs}
\usepackage{multirow}
\usepackage{tikz}
\usetikzlibrary{positioning}

\usepackage{subcaption}
\usepackage{bbold}
\usepackage{longtable}
\usepackage{slashbox}
\usepackage{tabularx}
\usepackage{mathrsfs}

\newcommand{\bb}{\color{black}}

\definecolor{Vred}{RGB}{251,128,114}
\definecolor{Gr}{RGB}{10,70,1}
\definecolor{Corr}{RGB}{128,239,128}
\definecolor{NOCorr}{RGB}{215,71,71}
\usepackage{colortbl}
\definecolor{lightblue}{RGB}{220,230,241}
\definecolor{lightgreen}{RGB}{226,240,217}
\definecolor{lightred}{RGB}{255,204,204}
\definecolor{mint}{RGB}{230,245,233}
\definecolor{peach}{RGB}{255,235,230}
\usepackage{algorithm}
\usepackage{algpseudocode}
\usepackage{bm}
\usepackage{placeins}
\usepackage{comment}
\usepackage{adjustbox}

\journal{IEEE TETCI}

\begin{document}
\begin{frontmatter}


%% Article title
\title{Sample-Aware Test-Time Adaptation for Medical Image-to-Image Translation}


\author[campus]{Irene Iele\textsuperscript{1}} %% Author name
\ead{irene.iele@unicampus.it}

\author[umea]{Francesco Di Feola\corref{cor1}\textsuperscript{1}} %% Author name
\ead{francesco.feola@umu.se}

\author[unige]{Matteo Tortora} %% Author name 
\ead{matteo.tortora@unige.it}

\author[unicamillus]{Rosa Sicilia} %% Author name \ead{francesco.feola@umu.se}

\author[unicamillus]{Valerio Guarrasi}
\ead{valerio.guarrasi@unicampus.it}

\author[campus,umea]{Paolo Soda}
\ead{p.soda@unicampus.it, paolo.soda@umu.se}

\affiliation[campus]{organization={Unit of Artificial Intelligence and Computer Systems, Department of Engineering,
Università Campus Bio-Medico di Roma},%Department and Organization
            %addressline={Via Álvaro del Portillo, 21}, 
            city={Rome},
            %postcode={00128}, 
            country={Italy}}

\affiliation[umea]{organization={Department of Diagnostics and Intervention, Radiation Physics, Biomedical Engineering, Umeå University},%Department and Organization
            city={Umeå},
            %postcode={901 87}, 
            country={Sweden}}
\affiliation[unige]{organization={Department of Naval, Electrical, Electronics and Telecommunications Engineering, University of Genoa},
%addressline={Via All'Opera Pia 11A},
city= {Genoa},
%postcode={16145},
country={Italy}}
\affiliation[unicamillus]{organization={UniCamillus-Saint Camillus International University of Health Sciences},%Department and Organization
            %addressline={Via Álvaro del Portillo, 21}, 
            city={Rome},
            %postcode={00128}, 
            country={Italy}}
            
\fntext[1]{These authors contributed equally to this work.}
\cortext[cor1]{Corresponding author.}
%\cortext[cor2]{Authors equally contributed to the work.}
%\thanks{\textsuperscript{1}These authors contributed equally to this work.}
%\textsuperscript{\dag}Authors equally contributed to the work.
%% Abstract

\begin{abstract}
Image-to-image translation has emerged as a powerful technique in medical imaging, enabling tasks such as image denoising and cross-modality conversion.
However, it suffers from limitations in handling out-of-distribution samples without causing performance degradation.
To address this limitation, we propose a novel Test-Time Adaptation (TTA) framework that dynamically adjusts the translation process based on the characteristics of each test sample.
Our method introduces a Reconstruction Module to quantify the domain shift and a Dynamic Adaptation Block that selectively modifies the internal features of a pretrained translation model to mitigate the shift without compromising the performance on in-distribution samples that do not require adaptation.
We evaluate our approach on two medical image-to-image translation tasks: low-dose CT denoising and $T_1$ to $T_2$ MRI translation, showing consistent improvements over both the baseline translation model without TTA and prior TTA methods.
Our analysis highlights the limitations of the state-of-the-art that uniformly apply the adaptation to both out-of-distribution and in-distribution samples, demonstrating that dynamic, sample-specific adjustment offers a promising path to improve model resilience in real-world scenarios.
The code is available at:
\sloppy\url{https://github.com/Sample-Aware-TTA/Code}
\end{abstract}

%% Keywords
\end{frontmatter}


\section{Introduction}
\label{section:intro}
Image-to-image translation converts an image from a source domain $X$ to a target domain $Y$ \cite{pang2021image}, and has found numerous applications across diverse fields, particularly in medical imaging \cite{kaji2019overview}, supporting different imaging modalities and key tasks such as noise reduction, image synthesis and super-resolution. 
In clinical settings, image-to-image translation may offer significant time and cost benefits, as it can reduce the need for repeated imaging exams on the same patient while minimizing exposure to high-dose radiation.
As a data-driven method, image-to-image translation is sensitive to distribution shift, which arises when the training and testing distributions differ due to variations in imaging systems, acquisition protocols, or anatomical structures \cite{hognon2024contrastive}.
Such shifts lead to out-of-distribution (OOD) data, samples that deviate from the training distribution, ultimately degrading model performance, limiting generalization, and reducing the quality of translated images.
To address this challenge, Test-Time Adaptation (TTA) has emerged as a strategy to adapt pretrained models to unseen data during inference without requiring retraining or access to large datasets, improving robustness in real-world scenarios, where institutions often lack the computational resources or data volume required to train models from scratch. 
For instance, TTA methods modify the input data, adjust intermediate feature representations, or update the model’s weights at inference time~\cite{liang2025comprehensive}. 
Despite recent advancements in this field, most existing methods apply adaptation strategies to all samples during inference~\cite{liang2024comprehensive}, whether they are OOD or in-distribution (ID) samples, that is data drawn from the same distribution as the training set.
However, this is a main limitation because TTA is currently designed to improve performance on OOD samples: when applied to ID data, it may disrupt the model’s optimal configuration and compromise its generalization~\cite{niu2022efficient}.
Therefore, striking the right balance between adapting to new data and preserving the original knowledge of the model is crucial to achieve consistent and reliable performance across diverse data distributions.
Moreover, while most studies have focused on TTA for predictive tasks~\cite{ma2022test, yang2022dltta, karani2021test, li2022self, valvano2021re, wen2024denoising, hu2021fully}, its application in the generative domain remains largely unexplored.

On these grounds, we hereby propose a sample-aware TTA method for medical image-to-image translation that dynamically adapts a pretrained translation model to OOD data while preserving its performance on ID data. 
To this end, we introduce a trainable reconstruction module that quantifies the degree of domain shift for each test input, and based on the estimated shift, an adaptation module applies feature-level transformations at multiple stages of the translation model, adjusting to the specific characteristics of each test sample.
We validate our approach through extensive experiments on two distinct tasks: Low Dose CT denoising (LDCT) and $T_1$ to $T_2$ MRI translation.

Hence, our contributions can be summarized as follows:
\begin{itemize}
    \item We propose a sample-aware TTA method that adapts a pre-trained image-to-image translation model to OOD samples, effectively narrowing the performance gap between ID and OOD samples while preserving ID performance.
    \item We design our TTA approach to be \textit{dynamic}, i.e., to be able to tailor the adaptation process to each OOD test sample and a strategy that allows  maximizing performance across varying degrees of distribution shift.
    \item We conduct extensive experiments to demonstrate the effectiveness of the proposed approach on two distinct medical image-to-image translation tasks, highlighting the task-agnostic nature of our method.
\end{itemize}
The rest of the paper is organized as follows: section~\ref{sec:sota} reviews related works on TTA, section~\ref{sec:methods} presents the methods, section~\ref{sec:experimental} describes the experimental configuration, section~\ref{sec:results} presents and discusses the results, whereas section~\ref{sec:conclusion} offers concluding remarks.

\section{Related Works}
\label{sec:sota}
TTA has emerged as a promising strategy to improve model robustness under distribution shifts~\cite{liang2025comprehensive}.
Early efforts in this direction date back to 2011, when Jain and Learned-Miller~\cite{jain2011online} investigated the use of Gaussian process regression to adapt a cascade of classifiers at inference time.
Since then, a variety of strategies have been proposed to adapt deep neural networks without requiring access to the training data or full model retraining~\cite{liang2025comprehensive, wang2024search}.

TTA is particularly relevant in medical imaging, where shifts frequently stem from differences in acquisition protocols, patient populations, and imaging devices.
Despite that, recent studies have primarily focused on using TTA to mitigate performance degradation in predictive tasks such as medical image classification~\cite{ma2022test, yang2022dltta}, and segmentation~\cite {karani2021test, li2022self, valvano2021re, wen2024denoising, hu2021fully}.
For example, Ma et al.~\cite{ma2022test} proposed a TTA strategy to address label distribution shifts by training multiple classifiers, each specialized for a class-dominated distribution. 
At test time, the outputs of these classifiers are dynamically combined and adapted to the test label distribution, using a consistency regularization loss to guide and calibrate their relative contributions. 
Yang et al.~\cite{yang2022dltta} introduced a method that dynamically adjusts the learning rate for each test sample based on the estimated distribution shift, enabling stable adaptation across both medical image classification and segmentation tasks.

In the context of medical semantic segmentation, Karani et al.~\cite{karani2021test} proposed input level adaptation by dynamically optimizing a shallow normalization network during inference. 
This network projects each test sample into a normalized space, guided by a denoising autoencoder prior trained to identify and correct implausible segmentations.
Building on this approach, Valvano et al.~\cite{valvano2021re} replaced the autoencoder prior with a discriminator network, trained to identify and provide corrective feedback to the normalization network at test time.
However, input-level adaptation alone may be insufficient to fully address complex domain shifts, as it cannot adapt deeper, task-relevant features.
To overcome this limitation, Li et al.~\cite{li2022self} extended TTA to both the input and feature levels, using a translation network and a multi-task segmentation network guided by an autoencoder-based reconstruction loss.
Another line of research focuses on test-time training using self-supervision~\cite{sun2020test}.
Instead of adapting intermediate features directly, the model weights are updated during inference guided by pre-text tasks such as image inpainting~\cite {pathak2016context} or rotation prediction~\cite{gidaris2018unsupervised}.
Building on this idea, Wen et al.~\cite{wen2024denoising} proposed an approach for CT image segmentation that employs a Y-shaped architecture based on a U-Net backbone.
The model includes a self-supervised denoising decoder that shares skip connections with the main segmentation branch.
It is jointly trained on segmentation and denoising tasks, while at inference time, only the encoder’s batch normalization layers are adapted using the denoising loss. 
Similarly, Hu et al.~\cite{hu2021fully} proposed a TTA method that updates only batch normalization parameters during inference, guided by two loss terms: one that promotes confident and diverse predictions across local image regions, and a contour regularization loss that encourages smooth boundaries to enhance spatial consistency in the segmentation outputs.
While TTA has been extensively studied for predictive tasks, its application to medical image-to-image translation has, to the best of our knowledge, been explored only in~\cite{he2021autoencoder}. 
In this work, similarly to~\cite{karani2021test}, He et al.~\cite{he2021autoencoder} proposed a self-supervised TTA approach for $T_1$ to $T_2$ MRI translation that  performs feature-level adaptation by using shallow adaptor networks, guided by an autoencoder-based reconstruction loss designed to capture domain shift.
However, the adaptation is applied uniformly across all test samples, without distinguishing between ID and OOD cases, which is a key limitation of the current literature. 
Furthermore, the method employs a fixed set of adaptors placed at both image and feature levels, inserted at predetermined locations within the task model, without accounting for the specific characteristics of each test sample, that can result in unnecessary transformations or suboptimal adaptation outcomes.

The review of the works presented in this section highlights the growing interest in TTA across various predictive domains. 
However, these methods have seen limited application in generative tasks, revealing a critical gap in the literature.
Moreover, the lack of selective, sample-specific TTA strategies for image-to-image translation underscores the need for approaches that activate adaptation only when necessary, preserving high fidelity and ensuring generalization across diverse medical imaging scenarios.

\section{Methods}
\label{sec:methods}
Let $\bm{x} \in \mathbb{R}^{w \times h}$ be a source image and $\bm{y} \in \mathbb{R}^{w \times h}$ its corresponding target image, where $w$ and $h$ denote its image width and height, respectively.
Image-to-image translation can be formulated as: 
\begin{equation}
    \bm{\hat{y}} = \mathcal{T}(\bm{x}) \simeq \bm{y}
\end{equation}
%with the goal of learning a mapping $\mathcal{T}$ that generates an output image $\bm{\hat{y}}$ as close as possible to the target image $\bm{y}$.
where $\mathcal{T}$ is a generic task model trained to map the source image $\bm{x}$ to its target counterpart $\hat{\bm{y}} \simeq \bm{y}$.
At inference, we apply TTA to refine the task model's performance on potentially OOD samples.
The goal is to reduce translation error by adapting the model to the individual test sample, such that:
\begin{equation}
\left| \hat{\bm{y}}^a - \bm{y} \right| \leq \left| \hat{\bm{y}} - \bm{y} \right|
\end{equation}
where $\hat{\bm{y}}^a$ denotes the synthesized output after applying TTA, and $|\cdot|$ is a generic error metric (e.g., $\ell_1$ or $\ell_2$ norm).
For the sake of presentation, Table~\ref{tab:algo_symbols} list the notation used in the manuscript.
\label{fig:Methods}
\begin{figure}[h!]
\centering
\includegraphics[width=0.75\textwidth]{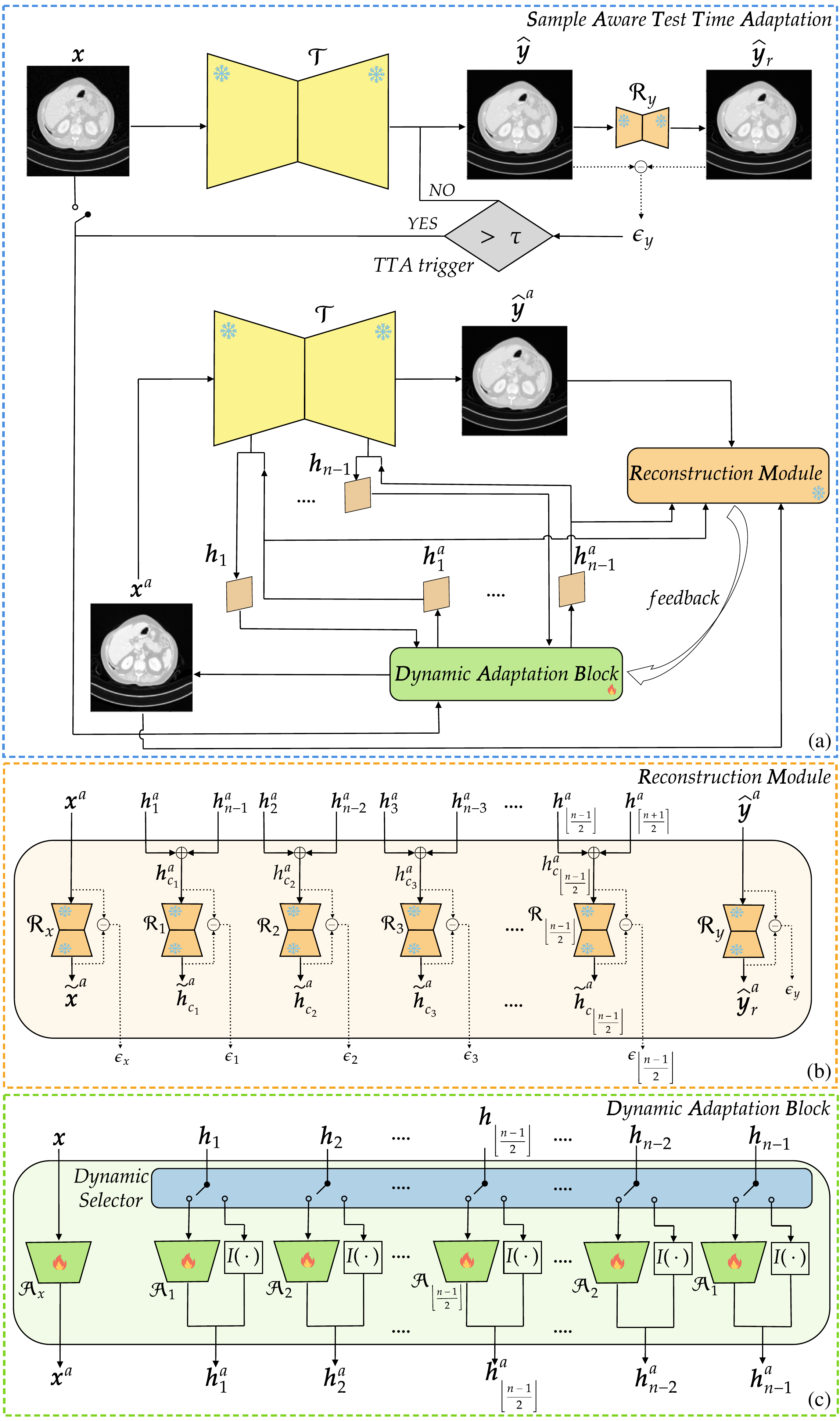}
%\captionsetup{justification=raggedright, singlelinecheck=false}
\caption[Schematic workflow of our Test-Time Adaptation approach]{Schematic workflow of our Test-Time Adaptation approach. (a) TTA phase, (b) Reconstruction Module, (c) Dynamic Adaptation Block, where a selector determines whether a feature map should be adapted or kept unchanged.}
\end{figure}
\FloatBarrier

\begin{table}[htpb]
\centering
\resizebox{0.8\textwidth}{!}{
\begin{tabular}{l p{10cm}}
\toprule
\textbf{Symbol} & \multicolumn{1}{c}{\textbf{Description}} \\
\midrule
$A_x$ & Input-level adaptor \\
$A_i$ & Set of intermediate adaptors \\
$\bm{h}_i$ & Feature maps of layer i\\
$\bm{h}^a_i$ & Adapted feature maps of layer i\\
$\bm{h}^{a}_{c_i}$ & Concatenation of encoder and decoder feature maps at level $i$, defined as $\bm{h}^{a}_{c_i} = \bm{h}^{a}_{i} \oplus \bm{h}^{a}_{n-i} $ \\
$\tilde{\bm{h}}^a_{ci}$ & Reconstruction of $\bm{h}^{a}_{c_i}$  \\
$M$ & Number of training steps used to update adaptors during TTA \\
$n$ & Number of layers in the task model \\
$\mathcal{T}$ & Task model used for Image-to-Image translation task \\
$\mathcal{T}^\omega$ & Task model with adaptors inserted at positions defined by $\omega$ \\
$\bm{x}$ & Source image \\
$\bm{x}^a$ & input after applying TTA\\
$\tilde{\bm{x}}^a$ & input produced by the reconstruction model $\mathcal{R}_y$ \\
$\bm{y}$ & Target image \\
$\hat{\bm{y}}$ & Synthetic target image \\
$\hat{\bm{y}}_r$ & Reconstructed output produced by the reconstruction model $\mathcal{R}_y$ \\
$\hat{\bm{y}}^a$ & Synthesized output of the full TTA pipeline\\
$\hat{\bm{y}}^a_r$ & Reconstruction of the synthesized output of the full TTA pipeline, obtained by passing $\hat{\bm{y}}^a$ through the output reconstruction module $\mathcal{R}_y$ \\
$\mathcal{R}_x$ & Input-domain reconstruction model \\
$\mathcal{R}_y$ & Output-domain reconstruction model\\
$\bm{\mathcal{R}}$ & Set all of intermediate reconstruction models, i.e., excluding input ($\mathcal{R}_x$) and output ($\mathcal{R}_y$) reconstruction models \\
$\mathcal{R}_i$ & $i$-th element of the set of intermediate reconstruction models $\bm{\mathcal{R}}$, which excludes the input ($\mathcal{R}_x$) and output ($\mathcal{R}_y$) models \\
$\mathcal{\epsilon}_x$ & Reconstruction error on the adapted input, i.e., $||\bm{x}^a - \tilde{\bm{x}}^a||$ \\
$\mathcal{\epsilon}_i$ & Reconstruction error on intermediate feature maps, computed at level $i$ \\
$\mathcal{\epsilon}_y$ & Reconstruction error on the output, i.e., $||\hat{\bm{y}} - \hat{\bm{y}}_r||$ \\
$\tau$ & Threshold for triggering adaptation \\
$\Omega$ & Search space of reconstruction model configurations \\
$\omega$ & A candidate configuration in $\Omega$ \\
\bottomrule
\end{tabular}
}
\caption{Summary of notation used throughout the manuscript sorted in alphabetical order.}
\label{tab:algo_symbols}
\end{table}
\FloatBarrier
Our TTA framework for medical image-to-image translation is illustrated in~\figurename~\ref{fig:Methods}.
Panel (a) shows that our approach comprises four main components:
the first is the pretrained task model $\mathcal{T}$, which remains frozen during adaptation.
The second is the TTA trigger, which selectively activates adaptation for OOD samples. 
The third is the Reconstruction Module (RM), shown in panel (b), which estimates domain shift at multiple feature levels and guides the adaptation process.
The fourth component is the Dynamic Adaptation Block (DAB), which transforms input features based on the shift estimated by the RM (panel (c)).
The key idea of our approach is to adapt only when necessary and in a sample-specific manner.
On the one hand, we retain the fixed parameters of $\mathcal{T}$ for ID samples, avoiding unnecessary adaptation.
On the other hand, we enable feature-level adaptation for OOD samples by dynamically selecting the most effective subset of feature adaptors.

To this end, we follow a multi-step training process.
First, we train the task model $\mathcal{T}$; then, we freeze $\mathcal{T}$ and train the RM on the same training set to learn feature distributions and quantify reconstruction errors at different levels of $\mathcal{T}$, serving as a proxy for domain shift during TTA. 
Following~\figurename~\ref{fig:Methods} (a), 
the test image $\bm{x}$ is first passed to the frozen task model $\mathcal{T}$, producing the output image $\hat{\bm{y}}$.
We then use the reconstructor $\mathcal{R}_y$ to compute the discrepancy between the task model’s output and its reconstruction:
\begin{equation}
\mathcal{\epsilon}_y = || \hat{\bm{y}}- \hat{\bm{y}}_r||, \quad \text{where} \quad \hat{\bm{y}}_r = \mathcal{R}_y(\hat{\bm{y}}).
\end{equation}
A trigger mechanism then activates TTA when the input image $\bm{x}$ is identified as OOD, i.e., when the reconstruction error $\mathcal{\epsilon}_y$ exceeds a threshold $\tau$.
Details on how we select the threshold $\tau$ are provided in section~\ref{subsec:experiments}.
Once TTA is triggered, the image $\bm{x}$ is processed by the DAB module, which uses the reconstruction error from the RM to perform input-level adaptation, producing an adapted input $\bm{x}^a$.
The adapted input $\bm{x}^a$ is then passed through the first layer of $\mathcal{T}$, producing a feature map $\bm{h}_1$.
The DAB transforms $\bm{h}_1$ into its adapted counterpart  $\bm{h}^a_1$, which replaces $\bm{h}_1$ and is propagated to the next layer of $\mathcal{T}$.
This iterative process of adaptation and forward propagation continues layer by layer, ultimately producing the final adapted output $\hat{\bm{y}}^a$.
In the following, section~\ref{Section: RB} and section~\ref{Section: DAB} detail the Reconstruction Module (RM) and the Dynamic Adaptation Block (DAB), respectively.

\subsection{Reconstruction Module (RM)}
\label{Section: RB} 
The Reconstruction Module (RM), shown in~\figurename~\ref{fig:Methods} (b), quantifies domain shift by computing reconstruction errors on both intermediate feature maps within the task model $\mathcal{T}$ and the generated output.
To this end, each layer of $\mathcal{T}$ is paired with a dedicated reconstruction network in RM, implemented as a convolutional autoencoder. 
These networks \( \{\mathcal{R}_i\}_{i=1 \ldots \lfloor \frac{n-1}{2} \rfloor} \), $\mathcal{R}_x$ and $\mathcal{R}_y$,
are independently trained on the same data used to train $\mathcal{T}$, learning the expected distribution of features at their respective levels. 
Once trained, they are kept frozen to serve as stable references for detecting distributional discrepancies. 
The core assumption is that the reconstruction error measured in inference by each reconstruction network is a proxy for domain shift, identifying deviations between test-time features and those observed during training.
At the highest resolution level, we use $\mathcal{R}_x$ and $\mathcal{R}_y$, to produce $\tilde{\bm{x}}^a$ and $\hat{\bm{y}}^a_r$, which are the reconstructions of the adapted input $\bm{x}^a$ and the task model’s output $\bm{y}^a$, respectively. The corresponding reconstruction errors are defined as:
\begin{equation}
    \mathcal{\epsilon}_x = \| \bm{x}^a - \tilde{\bm{x}}^a \|
\end{equation}
\begin{equation}
 \mathcal{\epsilon}_y = \| \hat{\bm{y}}^a - \hat{\bm{y}}^a_r \| \label{eq:l_y}
\end{equation}
where $\tilde{\bm{x}}^a = \mathcal{R}_x(\bm{x}^a)$ and $\hat{\bm{y}}^a_r  = \mathcal{R}_y(\hat{\bm{y}}^a)$ are the reconstructions produced by $\mathcal{R}_x$ and $\mathcal{R}_y$, respectively, and $||\cdot||$ denotes the $L_1$ distance.
At the intermediate feature levels, we employ the set of reconstruction models \( \{\mathcal{R}_i\}_{i=1 \ldots \lfloor \frac{n-1}{2}\rfloor} \), where $n$ denotes the total number of layers in $\mathcal{T}$.
At each level \( i \), we concatenate the encoder and decoder features, leveraging the symmetric structure of $\mathcal{T}$:
\begin{equation}
 \bm{h}^{a}_{c_i} = \bm{h}^{a}_{i} \oplus \bm{h}^{a}_{n-i}
\end{equation}
where  $\oplus$ denotes channel-wise concatenation.

The feature map $\bm{h}^{a}_{c_i}$ is then passed through the corresponding $\mathcal{R}_i$, yielding $\tilde{\bm{h}}^{a}_{c_i} = \mathcal{R}_i(\bm{h}^{a}_{c_i})$, with the corresponding reconstruction error defined as
\begin{equation}
\mathcal{\epsilon}_{\text{}_i} = \| \bm{h}^{a}_{c_i} - \tilde{\bm{h}}^{a}_{c_i}||.
\end{equation}
By fusing bottom-up and top-down information, each $\mathcal{R}_i$, gains a more complete representation of the feature space at level $i$, enhancing its ability to detect deviations indicative of domain shift.

% per VERSIONE PRECEDENTE vedi main+test_raw

\subsection{Dynamic Adaptation Block (DAB)}
\label{Section: DAB}
This block, illustrated in Figure~\ref{fig:Methods} (c), transforms the input image $\bm{x}$ and the features $\bm{h}_1, \bm{h}_2, ..., \bm{h}_n$, using the information provided by the RM to improve robustness to domain shift.
It is composed of trainable components, called adaptors, which perform feature-level transformations at multiple stages of the task model $\mathcal{T}$, and of a dynamic selector determining whether a given feature map should be adapted.
Let us remember that the adaptors, which are the only trainable component during TTA, are updated based on feedback from RM, with each reconstruction module $\mathcal{R}_i$ paired to a corresponding adaptor $A_i$ in DAB. 
To identify the most effective subset of adaptors, we propose a structured search framework that evaluates candidate combinations of reconstruction models within RM, using their feedback to guide the adaptation process and estimate domain shift.

Let $\bm{\mathcal{R}} = \{\mathcal{R}_1,\dots, \mathcal{R}_{\lfloor\frac{n-1}{2}\rfloor} \}$ be the set of intermediate reconstruction models. 
Note that $\bm{\mathcal{R}}$ explicitly excludes $\mathcal{R}_x$ and $\mathcal{R}_y$ from the search process:
on the one hand, $\mathcal{R}_x$ is fixed, as it operates directly on the adapted input image $\bm{x}^a=A_x(\bm{x})$, which always undergoes adaptation; on the other hand, $\mathcal{R}_y$, is consistently used to monitor the reconstruction error on the adapted output $\hat{\bm{y}}^a_r$.
%For these reasons, both $\mathcal{R}_x$ and $\mathcal{R}_y$ are therefore always employed and excluded from the search process, which only affects the intermediate reconstruction models in $\bm{\mathcal{R}}$.
%We define $\bm{\mathcal{R}_{\text{sel}}} \subseteq \bm{\mathcal{R}}$ as the subset of reconstruction models selected for TTA.
%This notation is consistently used across both the forward selection Algorithm~\ref{alg:Fs} and backward elimination Algorithm~\ref{alg:Bw} strategies to represent the evolving set of active reconstruction models during the search.
The search space is defined as:  
\begin{equation}
\Omega = \{ \omega \subseteq \Omega \mid \omega \neq \emptyset \} = \{\{\mathcal{R}_1\}, \dots, \{\mathcal{R}_1, \mathcal{R}_2\}, \dots, \bm{\mathcal{R}} \}
\end{equation}
where each $\omega \in \Omega$ represents a valid, non-empty subset of reconstruction models, considered as a candidate configuration for estimating domain shift during TTA.
Given a search strategy over the space $\Omega$, we evaluate each candidate configuration $\omega \in \Omega$ using equation~\ref{eq:l_y} to measure the reconstruction error of $\mathcal{R}_y$ over a limited number of adaptation steps $M$. 
At each step, we perform TTA to generate the adapted output $\hat{\bm{y}}^a=\mathcal{T}^\omega(\bm{x}^a)$, where $\mathcal{T}^\omega$ denotes the task model that performs inference while using the reconstruction models in $\omega$ to estimate domain shift and apply the corresponding adaptors. 
We then compute the reconstruction error $\mathcal{\epsilon}_y = \| \hat{\bm{y}}^a - \hat{\bm{y}}^a_r \|$, where $\bm{y}^a$ is the adapted output
and $\hat{\bm{y}}^a_r  = \mathcal{R}_y(\hat{\bm{y}}^a)$ is the corresponding reconstruction produced by the reconstruction model $\mathcal{R}_y$.
The optimal configuration $\omega^\ast$ is selected as the candidate that achieves the lowest reconstruction error over $M$ steps.

In details, we have an input-level adaptor $A_x$ and a set of intermediate adaptors $\{ A_i \}_{i=1, \ldots, \lfloor\frac{n-1}{2}\rfloor}$, where the number of intermediate adaptors corresponds to the first half of the architecture, as each $A_i$ is applied symmetrically to both encoder and decoder layers positioned at depth $i$ and $n-i$, respectively.
$A_x$ applies a sequence of convolutional layers preserving input image size.
$A_i$ is implemented as a $1\times1$ convolutional layer that transforms the feature maps without altering their spatial resolution.
A dynamic selector determines whether a given feature map should be adapted.
Based on its state, the intermediate feature $\bm{h}_i$ is either transformed by the adaptor, $\bm{h}_i^a=A_i(\bm{h}_i)$, or left unchanged as $\bm{h}_i^a = I(\bm{h}_i)=\bm{h}_i$, where $I(\cdot)$ denotes the identity operator.
%During TTA, adaptors are updated independently of one another.
During TTA, each adaptor is updated independently, allowing it to learn feature-level transformations based on the local characteristics of each test sample.
Each intermediate adaptor $A_i$ is applied both to its corresponding feature $\bm{h}_i$ and to its symmetric counterpart $\bm{h}_{n-i}$, located on the encoder and decoder branches of $\mathcal{T}$, respectively.
This design ensures consistent transformation across symmetric layers while minimizing the overall parameter overhead.

% per VERSIONE PRECEDENTE vedi main+test_raw
\begin{algorithm}[h]
\small
\caption{Dynamic Search}\label{alg:search}
\begin{algorithmic}[1]
\If {$\mathcal{\epsilon}_{\text{y}} > \tau $} 
\State $ \mathcal{\epsilon}^{\text{best}} \gets \infty$, $\omega^* \gets \text{None}$
    \ForAll {$\omega \in \Omega$} \Comment{Evaluate each configuration in the search space} 
        \State $ \mathcal{\epsilon}_{\text{steps}}^{\text{best}} \gets \infty$ \Comment{Initialize best error for the current configuration}
        \For {$i = 1$ to M} \Comment{Iterate over adaptor update steps}
            \State $\bm{y}^a = \mathcal{T}^\omega(\bm{x})$ \Comment{ Run the inference with the current configuration $\omega$}
            \State $\mathcal{\epsilon} \gets \mathcal{\epsilon}_y = \| \hat{\bm{y}}^a - \hat{\bm{y}}^a_r \| $, \Comment{Evaluate the output reconstruction error}
            \If {$\mathcal{\epsilon} < \mathcal{\epsilon}_{\text{steps}}^{\text{best}}$}
                \State $ \mathcal{\epsilon}_{\text{steps}}^{\text{best}} \gets \mathcal{\epsilon}$  \Comment{Update best error value for the}
                 \State \hfill \text{current configuration $\omega$}
            \EndIf
        \EndFor
        \If {$ \mathcal{\epsilon}_{\text{steps}}^{\text{best}}  <  \mathcal{\epsilon}^{\text{best}}$} \Comment{If the current best error value is better than}
          \State \hfill \text{the overall best error value}
            \State $ \mathcal{\epsilon}^{\text{best}} \gets \mathcal{\epsilon}_{\text{steps}}^{\text{best}} $, $\omega^* \gets \omega$ \Comment{Update best configuration $\omega^\ast{}$}
        \EndIf
    \EndFor
\EndIf
\State \Return $\omega^*$ \Comment{Return best configuration $\omega^\ast{}$}
\end{algorithmic}
\end{algorithm}

\FloatBarrier
% per VERSIONE PRECEDENTE algoritmi vedi main+test_raw

We propose an exhaustive grid search strategy as the core implementation of our sample-aware TTA framework. 
As detailed in Algorithm~\ref{alg:search}, the method evaluates all possible subsets of reconstruction modules in the configuration space $\Omega$ to identify the optimal feature-level adaptation for each test sample.
Adaptation is triggered only when the sample is detected as OOD, i.e., when the output reconstruction error $\mathcal{\epsilon}_y$ exceeds a fixed threshold $\tau$. 
For each candidate configuration $\omega \in \Omega$, the model performs $M$ update steps; the adapted output $\hat{\bm{y}}^a$ is reconstructed through $\mathcal{R}_y$ and the corresponding reconstruction error is computed. 
The configuration that achieves the lowest reconstruction error across all steps is selected as the optimal configuration $\omega^*$ and used for the final prediction. 
While computationally intensive, this exhaustive strategy enables precise, sample-specific adaptation and serves as the reference implementation throughout this work.

\section{Experimental Configuration}
\label{sec:experimental}
This section provides a comprehensive overview of the experimental setup used to evaluate the performance of our TTA approach.
We first describe the datasets used, including their characteristics and the preprocessing steps applied.
We then outline the experiments carried out, including the state-of-the-art competitor selected for comparison.
We describe implementation details, covering architectural choices, training configurations, and hyperparameter settings. 
Finally, we introduce the evaluation metrics used to quantitatively assess the effectiveness of our approach.

\subsection{Datasets}
We evaluate our approach on two image-to-image translation tasks: LDCT denoising and $T_{1}$ to $T_{2}$ MRI translation, whose main characteristics are summarized in Table~\ref{tab:dataset_summary}.
For the first task, we use the Mayo Clinic LDCT-and-Projection Data~\cite{moen2021low}, a public dataset which includes thoracic and abdominal high-dose CT (HDCT) scans and corresponding LDCT data simulated using a quarter-dose protocol.
From this dataset, we selected 60 patients comprising a total of 17,594 slices. 
Of these, 10 patients (5,936 slices) were used for training, while the remaining 50 patients (11,658 slices) were used for testing.
Preprocessing involved converting the raw DICOM files to Hounsfield Units (HU), selecting a display window centered at –400 HU with a width of 1400 HU, and normalizing all images to the range [–1, 1].

For the second task, we use the BraTS 2018 dataset~\cite{menze2014multimodal} that contains clinically pre-operative MRI scans acquired from multiple institutions, already preprocessed by the authors. 
We employed a total of 6,528 paired MRI slices, each consisting of a $T_1$- and a $T_2$-weighted image, amounting to 13,056 images in total. 
Of these, 5,760 pairs were used for training and 768 pairs for testing.
%The TTA phase is conducted on a test set consisting of 768 images from each domain, $T_{1}$ and $T_{2}$.
We additionally used the IXI dataset~\cite{IXI} as an external test set.
It includes paired $T_{1}$ and $T_{2}$ brain MRI scans collected from two clinical sites using different scanners: Hammersmith Hospital (Philips 3T system) and the Institute of Psychiatry using (Philips 1.5T system).
Following~\cite{he2021autoencoder}, the preprocessing included MNI space registration, white matter peak normalization, and volume resizing. 
A total of 70 3D scans were selected from the IXI dataset. From each scan, we extracted 21 axial slices, uniformly spaced between slice indices 120 and 180, using a 3 mm inter-slice distance. This procedure resulted in a final dataset of 2,800 2D slices.

It is worth noting that for the LDCT denoising task, no external validation set is used, as the test split of the Mayo Clinic dataset already offers sufficient variability to evaluate generalization.
With more than 11,000 test slices, OOD samples occur more frequently, making this setting particularly favorable for evaluating the effectiveness of sample-aware TTA.

\begin{table}[h]
\centering
\small
\resizebox{1\textwidth}{!}{
\begin{tabular}{lllllllm{10cm}}
\toprule
\textbf{Dataset} & \textbf{Domain source} & \textbf{Domain target} & \textbf{\# Instances} & \textbf{Train Set} & \textbf{Test Set} & \textbf{Image Size} \\
\midrule
Mayo Clinic LDCT \cite{moen2021low} & LDCT & HDCT & 17,594 slices & 5,936 slices & 11,658 slices & 512×512 \\
BraTS 2018 \cite{menze2014multimodal} & T1 MRI & T2 MRI & 6,528 slices & 5,760 slices & 768 slices & 240×240\\
IXI \cite{IXI} & T1 MRI & T2 MRI & 2,800 slices & — & 2,800 slices & 240×240 \\
\bottomrule
\end{tabular}
}
\caption{Summary of the datasets used in this study.} %Each row lists the source and target domains, the total number of instances, the training and testing set composition, and the image resolution.}
\label{tab:dataset_summary}
\end{table}

% per VERSIONE PRECEDENTE vedi main+test_raw

\subsection{Experiments} \label{subsec:experiments}

\begin{table}[h]
\centering
\small
\resizebox{1\textwidth}{!}{
\begin{tabular}{llm{10cm}}
\toprule
\textbf{Category} & \textbf{Experiment} & \textbf{Description} \\
\midrule
\multirow{3}{*}{Competitors} 
& No TTA & Baseline task model without TTA. \\
& He et al.~\cite{he2021autoencoder} & Static TTA approach from He et al.~\cite{he2021autoencoder} \\
&  He et al.\textsuperscript{*}  & Static TTA approach from He et al.~\cite{he2021autoencoder} applied at all feature levels of the task model $\mathcal{T}$.  \\
\midrule
\multirow{1}{*}{Our Approach} 
& $\mathrm{TTA}_{\mathrm{Grid}}$ & Sample-aware TTA with exhaustive grid search.  \\
\bottomrule
\end{tabular}
}
\caption{List of the proposed experiment configurations.}
\label{tab:experiment_ids}
\end{table}

As detailed in Table~\ref{tab:experiment_ids}, we evaluate our proposed and three competitors; furthermore we also investigated alternative search strategies for our TTA approach, widely discussed in section~\ref{Sec:Alternative Search Strategies}.
The first competitor, named No TTA, corresponds to the task model without adaptation and serves as a baseline for evaluating the benefits of our TTA strategy.
The second configuration correspond to the approach presented by He et al.~\cite{he2021autoencoder}.
It implements a static TTA method, and, to the best of our knowledge, is the only prior work addressing TTA in the context of image-to-image translation, as mentioned in section \ref{section:intro}.
The third configuration, named He et al.~\textsuperscript{*}, adopts the same adaptation strategy, integrating it directly into our task model $\mathcal{T}$ by inserting reconstruction models at all feature levels and applying them uniformly during inference, without any form of sample-specific selection.
This results in a full static adaptation strategy, applied uniformly across the test set, and serves as an ablation baseline for assessing the contribution of dynamic configuration selection.
As mentioned in section~\ref{section:intro}, applying TTA to ID samples can degrade performance by disrupting the task model’s optimal configuration learned during training. 

To mitigate this, we adopt a gating mechanism that triggers TTA only when it is likely to be beneficial, using a threshold $\tau$ that is based on  the reconstruction error produced by $\mathcal{R}_y$ on each test sample, as we describe in section~\ref{Section: DAB}. 
We set $\tau$ equal to  the $95^{th}$ percentiles of the reconstruction error distribution computed over the entire test set.
We deem that this choice is reasonable because samples with high reconstruction error are more likely to be OOD and, therefore, better candidates for adaptation. 
On the other hand, setting the threshold at the $95^{th}$ percentile is a conservative choice that limits unnecessary adaptation of  ID samples, thereby reducing the risk of performance degradation caused by overfitting or instability during test-time updates. 
This balance helps ensure that TTA is selectively applied to truly uncertain or anomalous cases, rather than introducing noise into confidently handled inputs.
Furthermore, to investigate the robustness of the  $\tau$ value used we performed a sensitivity analysis that evaluates $\tau$ multiple settings (i.e.,  $85^{th}$, $90^{th}$,  and $98^{th}$ percentiles of $\mathcal{R}_y$).
The results of this analysis, provided in~\ref{Appendix:Threshold Selection}, confirm that setting $\tau$ to the $95^{th}$ percentile is the best choice.

\subsection{Implementation details}
\label{Implementation details}
Our model consists of three main components: the task model, which performs the translation task and is implemented as a CycleGAN~\cite{zhu2017unpaired}; the reconstruction models, implemented as multi-level convolutional autoencoders; 
and the adaptors, implemented as $1 \times 1$ convolutional layers.
For more details on the model's components, please refer to~\ref{appendix:Preliminaries}. 
We trained the task models $\mathcal{T}$  using the Adam Optimizer with an initial learning rate of $2 \times 10^{-4}$ and a batch size of 8. 
The training schedule consists of 50 epochs with a fixed learning rate, followed by 50 epochs during which the learning rate decays linearly to zero, for a total of 100 epochs. 
This strategy balances the need for sufficient iterations to capture complex patterns while mitigating overfitting.
The learning rate decay stabilizes convergence in the final stages of training, and the chosen batch size reflects a trade-off between memory constraints and training stability.

Each reconstruction model $\mathcal{R}$ was trained independently using Adam optimizer with the initial learning rate of $1 \times 10^{-3}$, and a batch size of 8. 
The training schedule consists of 20 epochs with a fixed learning rate, followed by 80 epochs during which the learning rate decays linearly to zero, for a total of 100 epochs. 
This setup ensures consistency in training dynamics across all reconstruction models while enabling each of them to specialize on different features extracted from the task model.

All experiments were conducted on a high-performance computing cluster, equipped with one GPU NVIDIA A100, optimized for large-scale deep learning workloads.

\subsection{Evaluation metrics}
\label{Evaluation metrics}
We used three well-established image quality assessment metrics to quantify the quality of the generated images. 
They are the Mean Absolute Error ($\mathrm{MAE}$), Peak-Signal-to-Noise Ratio ($\mathrm{PSNR}$), and Structural Similarity Index ($\mathrm{SSIM}$) \cite{di2023comparative}.

The $\mathrm{MAE}$  measures the absolute difference between the pixel values of the generated image $\hat{y}$ and the target image $y$:
\begin{equation}\label{eq:mse}
\mathrm{MAE}(\hat{y}, y) = \frac{1}{mn}\sum_{i=0}^{m-1}\sum_{j=0}^{n-1}|\hat{y}_{i,j} - y_{i,j}|
\end{equation}
where $m$ and $n$ are the number of rows and columns in the images, respectively, and $y_{i,j}$, $\hat{y}_{i,j}$ denotes the pixel elements at the $i$-th row and $j$-th column of $y$ and $\hat{y}$, respectively. It varies in the range $[0, +\infty]$, with a lower $\mathrm{MAE}$ indicating a better match between $y$ and $\hat{y}$.

The $\mathrm{PSNR}$ compares the maximum intensity in the generated image ($\mathrm{MAX}_{\hat{y}}$) with the error between the generated image $\hat{y}$ and the target image $y$ given by the mean squared error ($\mathrm{MSE}$):
\begin{equation}\label{eq:psnr}
\mathrm{PSNR}(\hat{y}, y) = 10 \times \log_{10}\left(\frac{\mathrm{MAX}_{\hat{y}}^2}{\mathrm{MSE}(\hat{y}, y)}\right)
\end{equation}
It varies in the range $[0, +\infty]$, where higher  $\mathrm{PSNR}$ values indicate better quality. %while lower values may suggest more distortions or errors in $\hat{y}$.

The SSIM~\cite{wang2004image} computes the similarity between two images as a function of luminance, contrast, and structure:
\begin{equation}\label{eq:ssim}
\mathrm{SSIM}(\hat{y}, y) = 
\frac{(2\mu_{\hat{y}}\mu_{y} + c_{1})(2\sigma_{\hat{y}y} + c_{2})}
{(\mu_{\hat{y}}^2 + \mu_{y}^2 + c_{1})(\sigma_{\hat{y}}^2 + \sigma_{y}^2 + c_{2})}
\end{equation}
where $\mu_{\hat{y}}$, $\mu_{y}$ are mean intensities of the pixels in $\hat{y}$ and $y$, respectively; $\sigma_{\hat{y}}^2$ and $\sigma_{y}^2$ are the variances, $\sigma_{\hat{y}y}$ is the covariance whilst $c_{1}$ and $c_{2}$ are constant values to avoid numerical instabilities. The ($\mathrm{SSIM}$) varies in the range $ [0,1] $: the higher its value, the greater the similarity between the two images.

\section{Results and Discussion}
\label{sec:results} 
This section presents a comprehensive evaluation of our sample-aware TTA approach through both quantitative and qualitative analyses across two medical image translation tasks; LDCT denoising and $T_1$ to $T_2$ MRI translation. 
We first compare the performance of our method against the competitors described in section~\ref{subsec:experiments}, followed by a detailed analysis of their core design assumptions and limitations in addressing domain shifts at test time.
Next, we presents a visual comparison of the translated outputs, highlighting the qualitative improvements achieved by our method in terms of anatomical fidelity and noise suppression.
section~\ref{Sec:Alternative Search Strategies} evaluates alternative search strategies for identifying optimal adaptor configurations, and section~\ref{sec: Computational analysis} examines the computational costs associated with each search strategy, emphasizing trade-offs between inference cost and adaptation effectiveness.

\label{Quantitative evaluation of TTA}
Table~\ref{tab:tasks} summarizes the experimental results, and is organized into three sections according to the translation task: LDCT denoising, $T_1$-to-$T_2$ MRI translation on the BraTS 2018 dataset, and $T_1$-to-$T_2$ MRI translation on the IXI dataset.
For each task, there are four rows corresponding to the configurations presented in section~\ref{subsec:experiments}.
By column, each metric in table is divided in two sub-columns $\bm{A}$ and $\bm{B}$.
Here, $\bm{A}$ denotes the full test set, while $\bm{B} \subset \bm{A}$ includes only the samples identified as OOD by our method.
OOD samples are selected using a selection threshold $\tau$ set to 95$^{th}$ percentile of the reconstruction error distribution computed over the entire test set, as detailed in section~\ref{subsec:experiments}.
In each section, the best-performing results are highlighted in green and demonstrate statistically significant differences from the others, satisfying the Wilcoxon test with Bonferroni correction~\cite{bonferroni1936teoria}.
Although the competitors do not support OOD sample identification, we report their performance on $\bm{B}$ to enable a comprehensive comparison under OOD conditions. 

% per VERSIONE PRECEDENTE vedi main+test_raw

\begin{table*}[!h]
\centering
%\begin{adjustbox}{center}
%\resizebox{16.5cm}{!}{
\resizebox{1\textwidth}{!}{
\begin{tabular}{l|ll|cc|cc|cc}
\toprule
\textbf{Task} & & \hspace{-0.3cm}\textbf{Experiment} & \multicolumn{2}{c|}{\textbf{SSIM $\uparrow$}} 
& \multicolumn{2}{c|}{\textbf{MAE $\downarrow$}} 
& \multicolumn{2}{c}{\textbf{PSNR $\uparrow$}} \\
\toprule
 & \textbf{} & \textbf{} 
& \textbf{$\bm A$} & $\bm{B \subset A}$ 
& \textbf{$\bm A$} & $\bm{B \subset A}$ 
& \textbf{$\bm A$} & $\bm{B \subset A}$ \\
\multirow{4}{4em}{LDCT \\ Denoising}& \multirow{3}{4em}{Competitors}& \hspace{0.5cm} No TTA & .693 \textsuperscript{$\pm$ .199}  & .647 \textsuperscript{$\pm$ .243}$^\dagger$ & .063 \textsuperscript{$\pm$ .048} &  .118 \textsuperscript{$\pm$ .071}$^\dagger$  & 27.913 \textsuperscript{$\pm$ 5.793} & 18.064 \textsuperscript{$\pm$ 2.164}$^\dagger$ \\
& & \hspace{0.5cm} He et al.~\cite{he2021autoencoder} & .384 \textsuperscript{$\pm$ .167}  & .552 \textsuperscript{$\pm$ .902}$^\dagger$ & .388 \textsuperscript{$\pm$ .105} &  .331 \textsuperscript{$\pm$ .057}$^\dagger$  & 15.723 \textsuperscript{$\pm$ 1.524} & 16.388 \textsuperscript{$\pm$ 1.183}$^\dagger$ \\
& & \hspace{0.5cm} He et al.* & .645 \textsuperscript{$\pm$ .211}  & .729 \textsuperscript{$\pm$ .273}$^\dagger$ & .089 \textsuperscript{$\pm$ .069} &  .084 \textsuperscript{$\pm$ .062}$^\dagger$  & 24.875 \textsuperscript{$\pm$ 4.277} & 25.557 \textsuperscript{$\pm$ 5.670}$^\dagger$ \\
\cmidrule{2-9}
& \multirow{1}{*}{Our approach}& \hspace{0.5cm}  $\mathrm{TTA}_{\mathrm{Grid}}$ &\cellcolor{lightgreen} .699 \textsuperscript{$\pm$ .203} & \cellcolor{lightgreen} .769 \textsuperscript{$\pm$ .289} & \cellcolor{lightgreen} .060 \textsuperscript{$\pm$ .045} & \cellcolor{lightgreen} .063 \textsuperscript{$\pm$ .053} & \cellcolor{lightgreen} 28.464 \textsuperscript{$\pm$ 5.492} & \cellcolor{lightgreen} {29.204 \textsuperscript{$\pm$ 6.137}} \\ 
\midrule\midrule
\multirow{4}{4em}{MRI $T_1$-$T_2$ \\(BraTS 2018)}& \multirow{3}{4em}{Competitors}& \hspace{0.5cm} No TTA &  \cellcolor{lightgreen}.858\textsuperscript{$\pm$ .040} &\cellcolor{lightgreen}.866 \textsuperscript{$\pm$ .012}$^\dagger$  &\cellcolor{lightgreen} .037 \textsuperscript{$\pm$ .010} &\cellcolor{lightgreen} .039 \textsuperscript{$\pm$ .005}$^\dagger$  & \cellcolor{lightgreen} 26.459 \textsuperscript{$\pm$ 1.864} &  \cellcolor{lightgreen} 25.241 \textsuperscript{$\pm$ .681}$^\dagger$ \\
& & \hspace{0.5cm} He et al.~\cite{he2021autoencoder} & .330 \textsuperscript{$\pm$ .519} & .279 \textsuperscript{$\pm$ .031}$^\dagger$ & .571 \textsuperscript{$\pm$ .060} & .591 \textsuperscript{$\pm$ .059}$^\dagger$ & 7.23 \textsuperscript{$\pm$ .899} & 7.701 \textsuperscript{$\pm$ .844}$^\dagger$\\
& & \hspace{0.5cm} He et al.* & .803 \textsuperscript{$\pm$.045} & .795 \textsuperscript{$\pm$ .024}$^\dagger$  & .079 \textsuperscript{$\pm$ .028} & .075 \textsuperscript{$\pm$ .014}$^\dagger$  & 20.498 \textsuperscript{$\pm$2.760} & 19.403 \textsuperscript{$\pm$ 1.673}$^\dagger$  \\
\cmidrule{2-9}
& \multirow{1}{*}{Our approach}& \hspace{0.5cm}  $\mathrm{TTA}_{\mathrm{Grid}}$ & .855 \textsuperscript{$\pm$ .042} & .800 \textsuperscript{$\pm$ .030} & .039 \textsuperscript{$\pm$ .013} & .067 \textsuperscript{$\pm$ .018} & 26.239 \textsuperscript{$\pm$ 2.293} & 20.557 \textsuperscript{$\pm$ 2.517} \\ 
\midrule\midrule
\multirow{4}{4em}{MRI $T_1$-$T_2$ \\(IXI)}& \multirow{3}{4em}{Competitors}& \hspace{0.5cm} No TTA &  .692 \textsuperscript{$\pm$ .046} & .700 \textsuperscript{$\pm$.048}$^\dagger$ & .115 \textsuperscript{$\pm$ .022} & .126 \textsuperscript{$\pm$ .018}$^\dagger$ & 18.622 \textsuperscript{$\pm$ 1.347} & 17.953 \textsuperscript{$\pm$ 1.044}$^\dagger$ \\
& & \hspace{0.5cm} He et al.~\cite{he2021autoencoder} & .289 \textsuperscript{$\pm$ .115} & .271 \textsuperscript{$\pm$ .114}$^\dagger$ & .481 \textsuperscript{$\pm$ .123} & .504 \textsuperscript{$\pm$ .115}$^\dagger$ & 9.851 \textsuperscript{$\pm$ 2.018} &  9.470 \textsuperscript{$\pm$ 1.536}$^\dagger$ \\
& & \hspace{0.5cm} He et al.* & .704 \textsuperscript{$\pm$ .045} & .724 \textsuperscript{$\pm$ .044}$^\dagger$ & .103 \textsuperscript{$\pm$ .017} & .090 \textsuperscript{$\pm$ .018}$^\dagger$ & 18.410 \textsuperscript{$\pm$1.996} & 18.477 \textsuperscript{$\pm$ 2.259}$^\dagger$ \\
\cmidrule{2-9}
& \multirow{1}{*}{Our approach}& \hspace{0.5cm}  $\mathrm{TTA}_{\mathrm{Grid}}$ &\cellcolor{lightgreen} .729 \textsuperscript{$\pm$ .068} & \cellcolor{lightgreen} .761 \textsuperscript{$\pm$ .064} &  \cellcolor{lightgreen}.093 \textsuperscript{$\pm$ .021} &  \cellcolor{lightgreen}.090 \textsuperscript{$\pm$ .023} & \cellcolor{lightgreen} 19.951 \textsuperscript{$\pm$ 1.842} & \cellcolor{lightgreen} 20.317 \textsuperscript{$\pm$ 2.179}  \\ 
\bottomrule
\end{tabular}
}
%\end{adjustbox}
\caption{Quantitative comparison of different experiments across different translation tasks. 
Metrics are reported for the entire test set $\bm{A}$ and for the subset $\bm{B} \subset \bm{A}$. 
The best results for each task are highlighted in green.
The dagger symbol ($^\dagger$) denotes results from configurations that do not support OOD sample identification but are evaluated on the subset $\bm{B} \subset \bm{A}$, identified by our approach, to illustrate how these configurations would perform on the detected OOD samples.
} 
%The best results are marked in bold and their statistical significance is assessed with Wilcoxon signed rank test ($p < 0.05$).}
\label{tab:tasks}
\end{table*}

In the LDCT denoising task, $\mathrm{TTA}_{Grid}$ achieves the highest performance across all three metrics on both the full test set $\bm{A}$ and the OOD subset $\bm{B} \subset \bm{A}$, clearly outperforming all competitors. 
The performance improvement is evident on $\bm{B}$, where domain shift is more severe, which is expected since $\mathrm{TTA}_{Grid}$ dynamically tailors the adaptation on a per-sample basis, rather than applying a fixed adaptation across all test samples.
The comparison with the static method proposed by He et al.\textsuperscript{*}, further underscores the importance of the adaptive nature of our approach.
While He et al.\textsuperscript{*} provides modest gains over the task model without adaptation, No TTA, it lacks the flexibility to adjust to the specific characteristics of individual test samples.
These findings demonstrate that $\mathrm{TTA}_{\mathrm{Grid}}$ not only improves average denoising but also enhances robustness under domain shift, a critical factor in clinical scenarios characterized by low-dose protocols and heterogeneous image quality.

Turning to the results for $T_1$-to-$T_2$ MRI translation on the BraTS 2018 dataset, $\mathrm{TTA}_{Grid}$ does not provide measurable benefits.
Our method achieves comparable performance to the task model without adaptation (No TTA) on  $\bm{A}$ and even shows a slight performance drop on the OOD subset $\bm{B}$. 
This outcome is expected, as both the training and test samples are drawn from the same distribution; since the task model already generalizes effectively, TTA provides limited benefit and can even degrade performance. 
These findings underscore the importance of first assessing the presence of domain shift before applying TTA, highlighting the limitations of indiscriminate adaptation in ID scenarios.

To further investigate this hypothesis, we assess the performance of the task model on the IXI dataset, used here as an external test set.
In this more realistic setting, where the dataset introduces a substantial domain shift due to differences in scanner types and acquisition protocols, $TTA_{grid}$ achieves consistent performance gains across all metrics, with notable improvement on the OOD subset $\bm{B}$. 
He et al.~\cite{he2021autoencoder} achieves the poorest performance, failing to improve translation quality in either task.
This may be due to the uniform application of adaptation via a static design: feature-level adaptation is carried out through a fixed set of adaptors inserted at all layers of the task model $\mathcal{T}$, without accounting for input-specific characteristics or supporting dynamic configuration.
Additionally, the reconstruction models are trained jointly using a single cumulative loss function, which limits their ability to specialize and reduces their effectiveness in capturing domain shifts.
As a result, the adaptors are less capable of applying appropriate feature transformations at different levels of the task model $\mathcal{T}$.
These shortcomings underscore the limitations of a one-size-fits-all adaptation strategy that treats all test samples equally, regardless of whether they are ID or OOD.
He et al.\textsuperscript{*} achieve moderately better performance by training each reconstruction model independently, allowing them to capture feature-level variations more effectively and serve as a stronger proxy for measuring domain shift. However, the adaptation mechanism remains static and applied uniformly across all test samples, limiting the adaptors’ ability to tailor feature-level transformations to input-specific characteristics.
Although both He et al.~\cite{he2021autoencoder} and He et al.\textsuperscript{*} do not distinguish between ID and OOD samples, we  report their performance on the OOD subset $\bm{B}$, as identified by $\mathrm{TTA}_{\mathrm{Grid}}$.
He et al.\textsuperscript{*} attains suboptimal results on the full test set $\bm{A}$, confirming that TTA is most effective when selectively applied, but it also fails to improve translation performance on the OOD subset $\bm{B}$.
On the subset $\bm{B}$, we further observe that, even when focusing only on OOD samples, fixed reconstruction strategies remain less effective than our approach.
This comparison reinforces the importance of tailoring the adaptation to the specific distributional properties of each test sample.

\begin{figure}[ht]
    \centering
    \begin{subfigure}[b]{0.3\linewidth}
        \includegraphics[width=\linewidth]{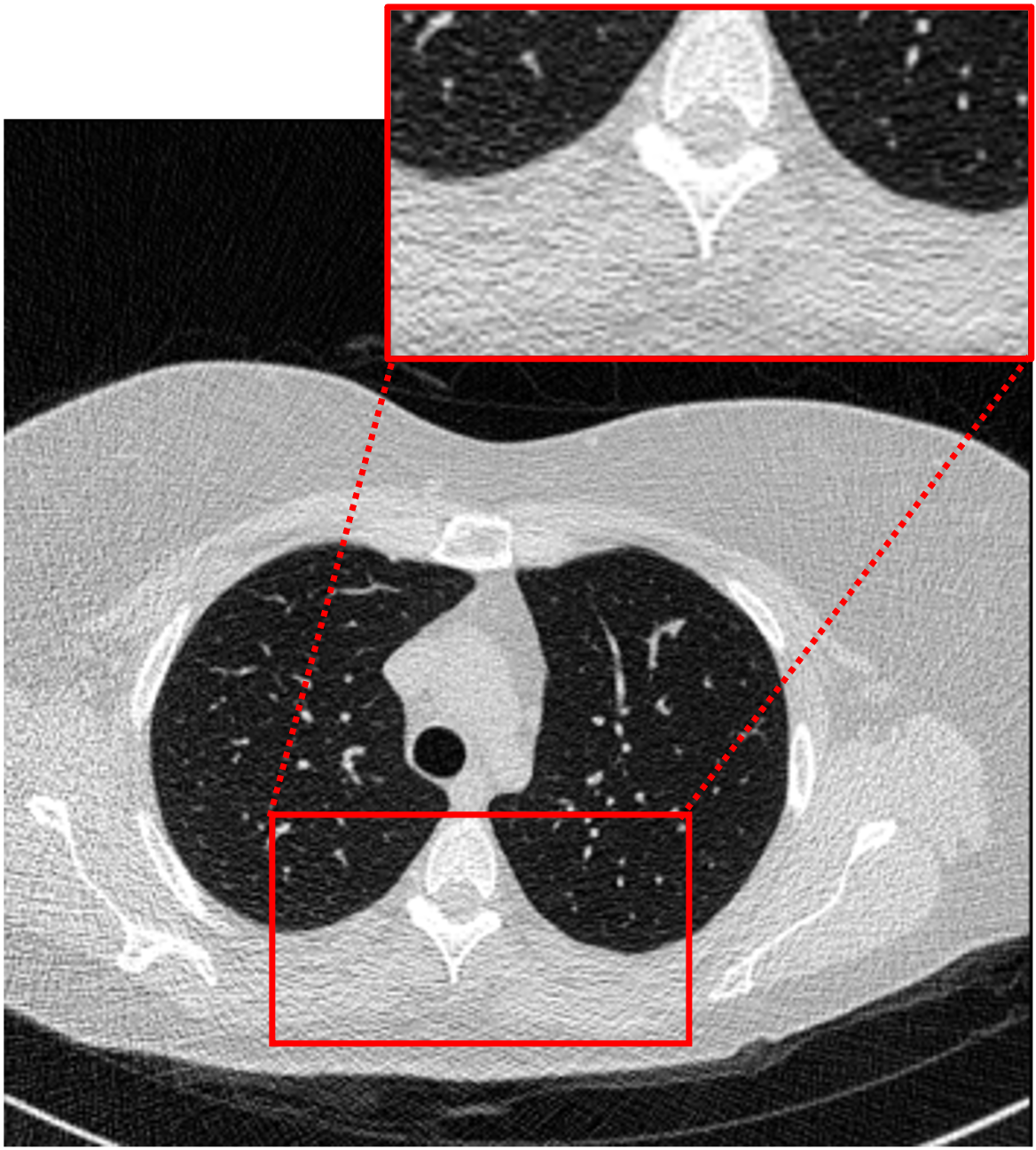}
        \caption{Low-dose input}
    \end{subfigure}
     \hspace{0.02\linewidth}
    \begin{subfigure}[b]{0.3\linewidth}
        \includegraphics[width=\linewidth]{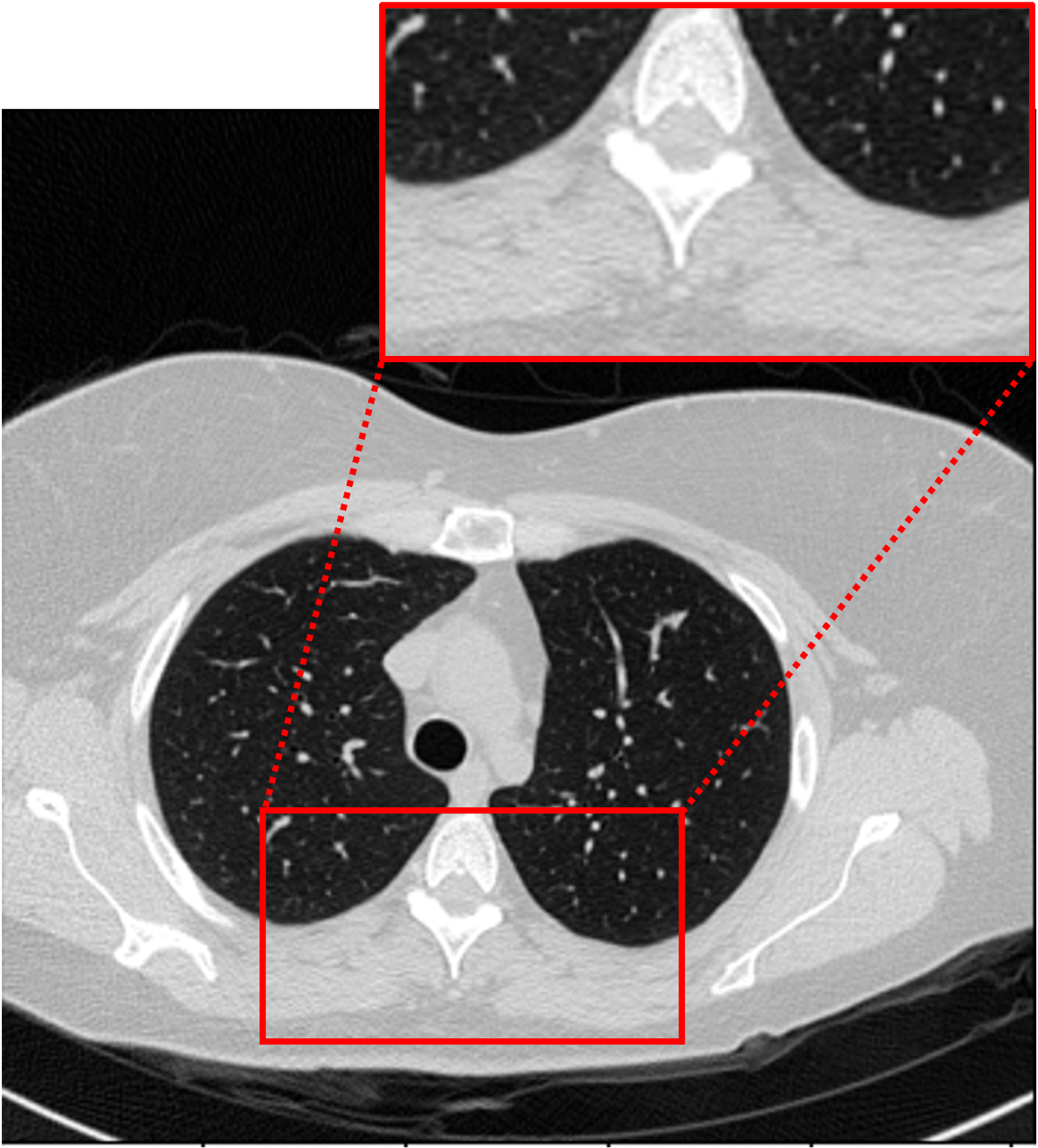}
        \caption{high-dose reference}
    \end{subfigure}
    \hspace{0.02\linewidth}
    \begin{subfigure}[b]{0.3\linewidth}
        \includegraphics[width=\linewidth]{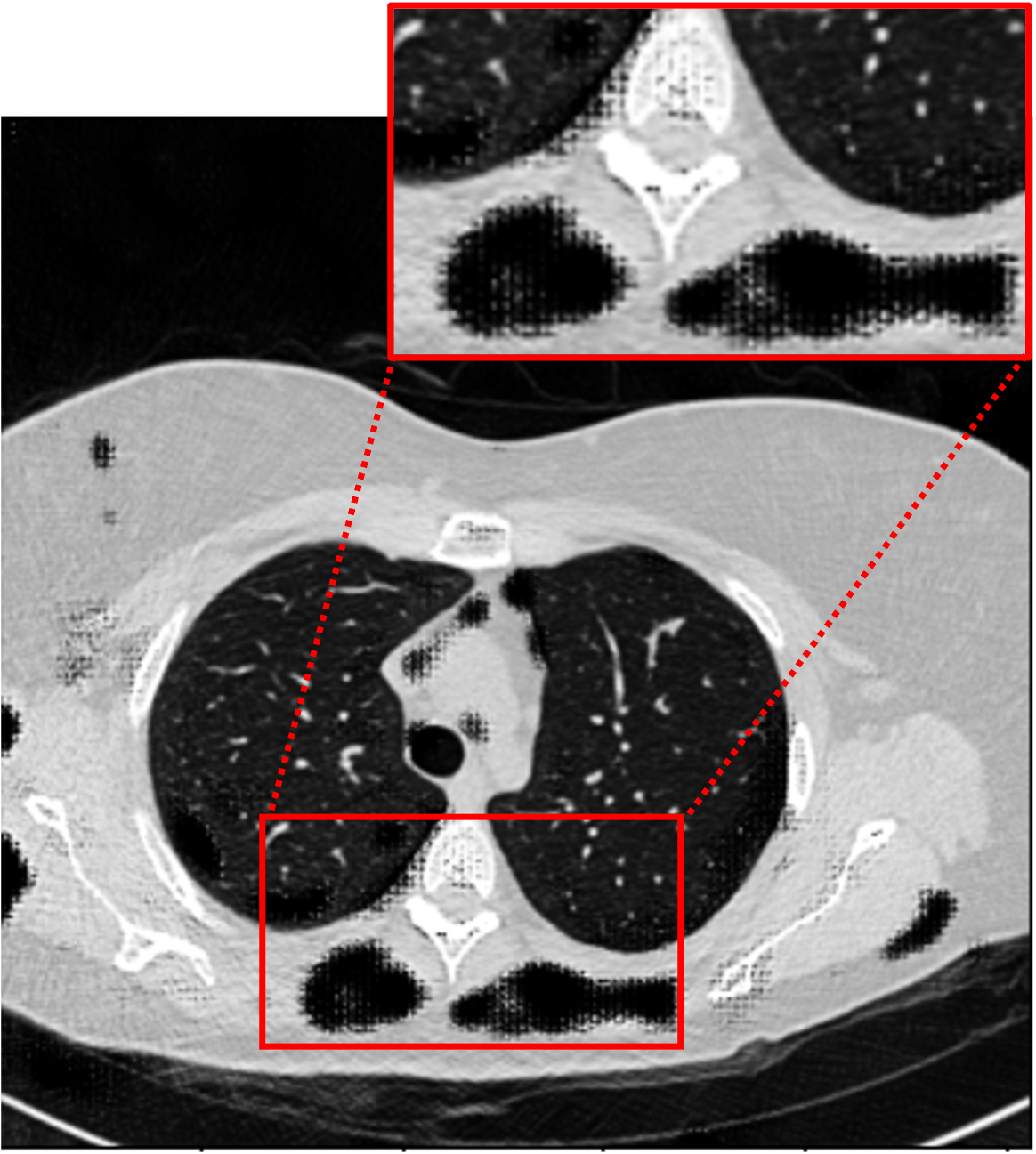}
        \caption{No TTA}
    \end{subfigure}
    
    \vspace{0.4cm}
    
    \begin{subfigure}[b]{0.3\linewidth}
        \includegraphics[width=\linewidth]{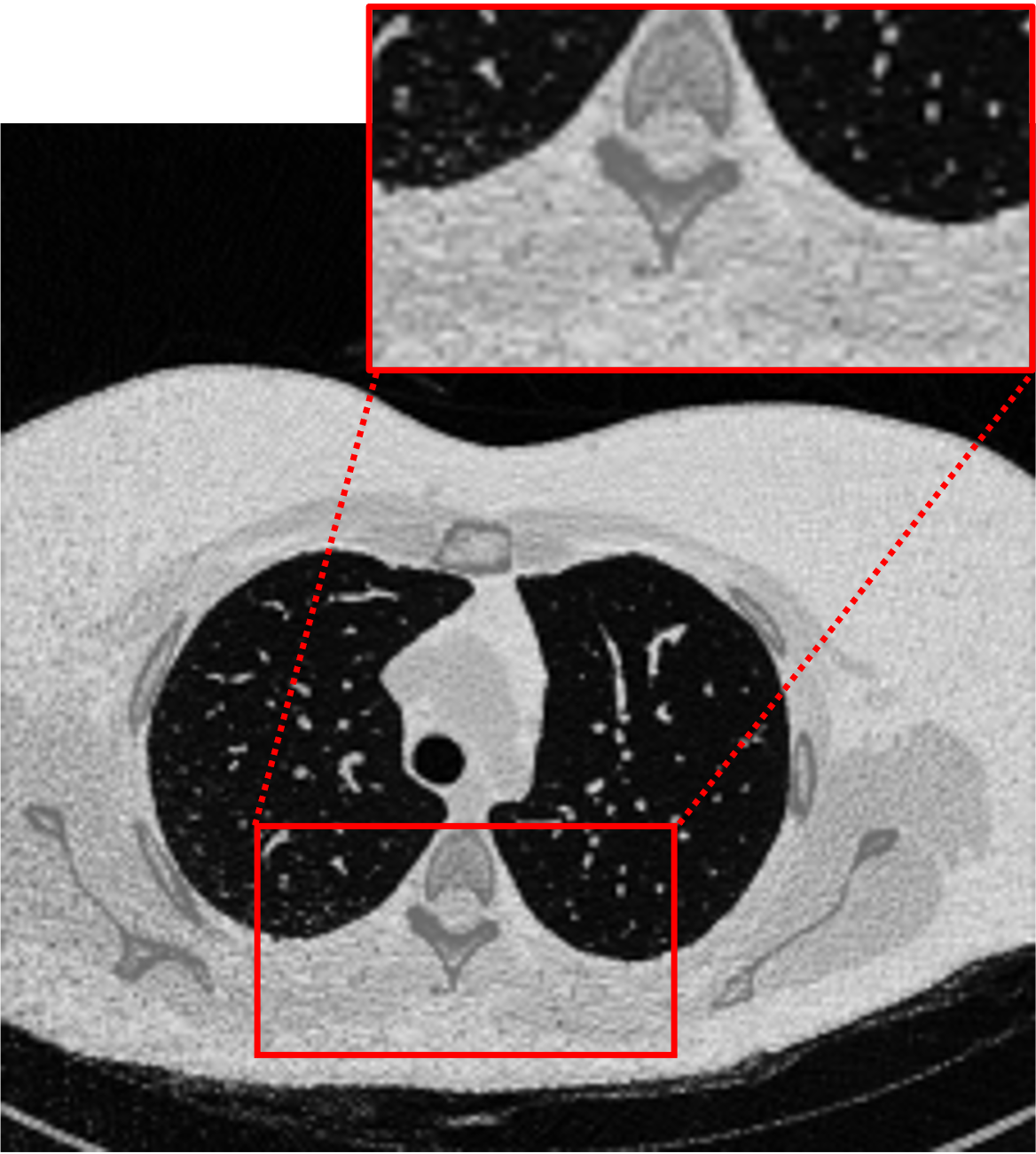}
        \caption{He et al.~\cite{he2021autoencoder}}
    \end{subfigure}
     \hspace{0.02\linewidth}
    \begin{subfigure}[b]{0.3\linewidth}
        \includegraphics[width=\linewidth]{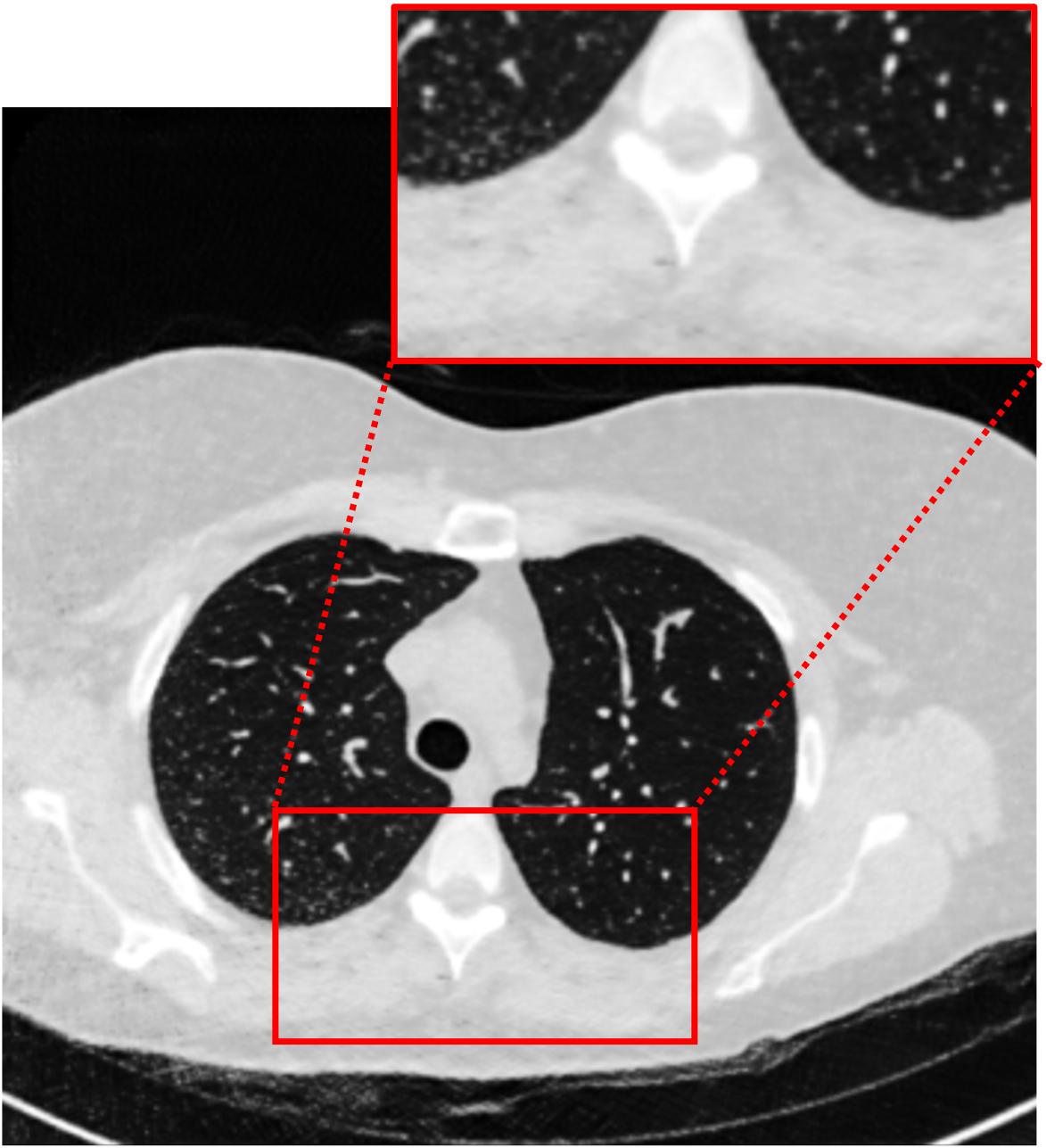}
        \caption{He et al.\textsuperscript{*}}
    \end{subfigure}
     \hspace{0.02\linewidth}
    \begin{subfigure}[b]{0.3\linewidth}
        \includegraphics[width=\linewidth]{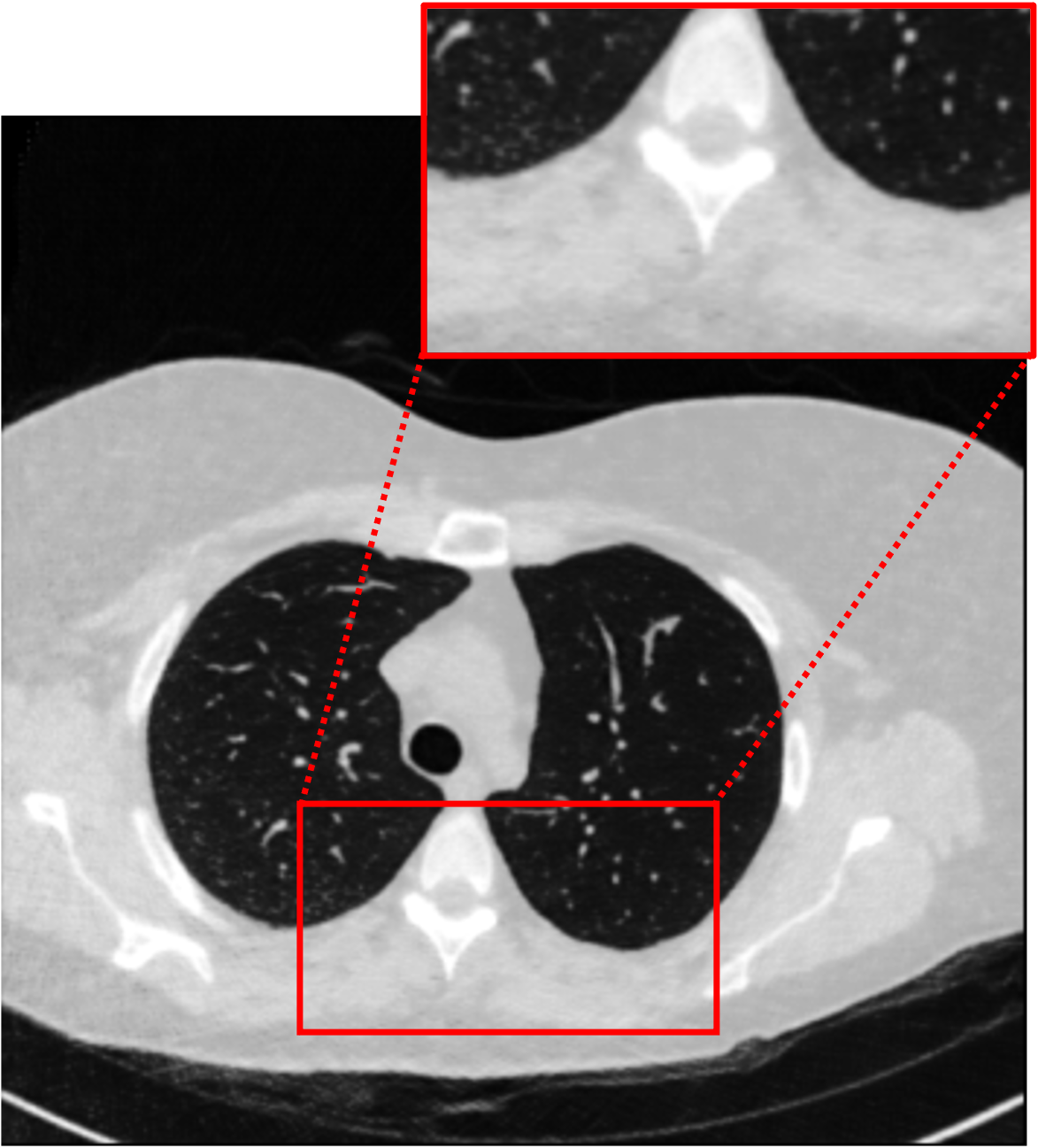}
        \caption{$\mathrm{TTA}_{\mathrm{Grid}}$}
    \end{subfigure}

    \caption{
Visual comparison of denoised CT slices. Zoomed ROIs highlight key regions where TTA yields notable improvements in denoising quality.}
    \label{fig_LDCT}
\end{figure}

\begin{figure}[ht]
    \centering
    \begin{subfigure}[b]{0.3\linewidth}
        \includegraphics[width=\linewidth]{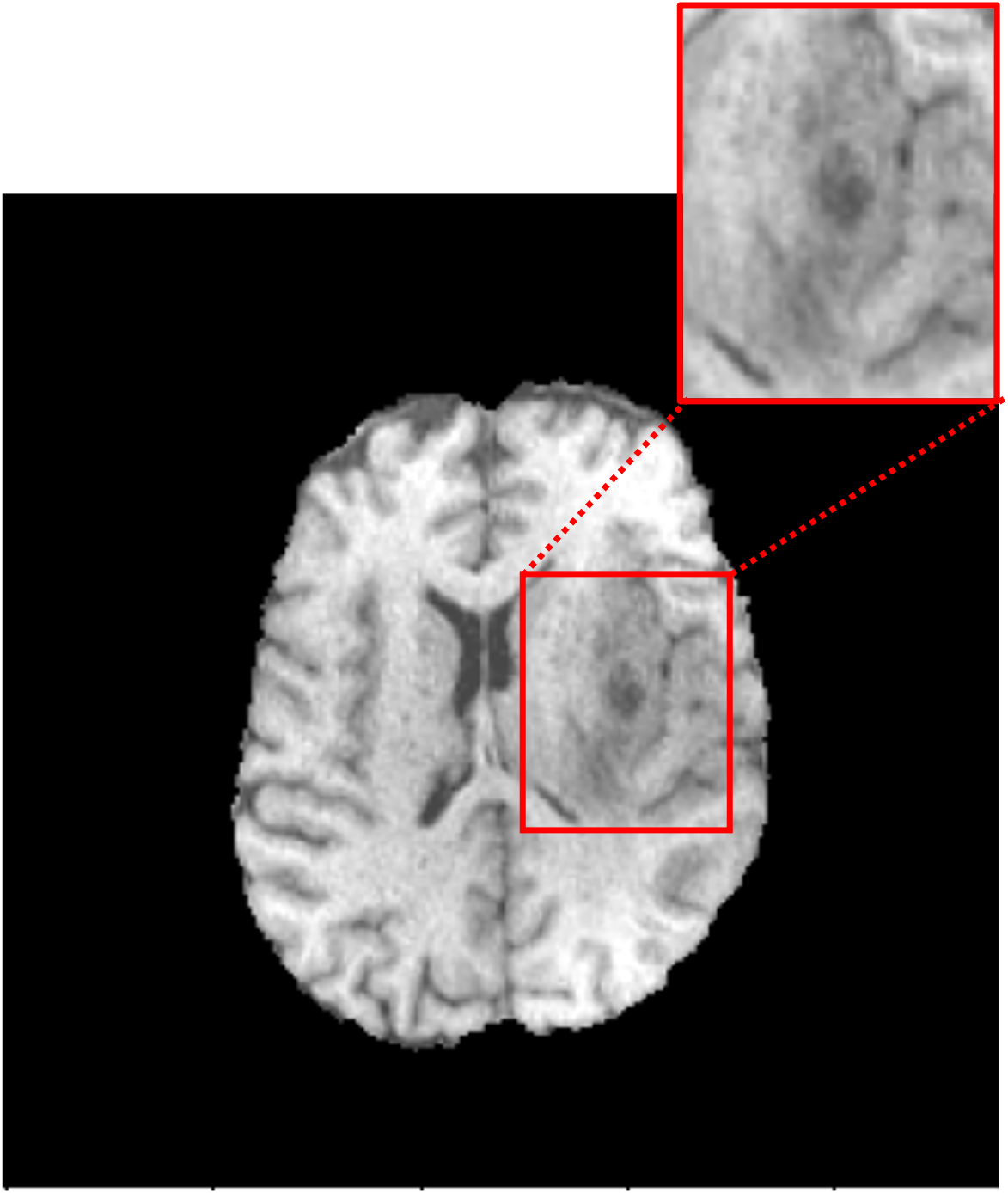}
        \caption{$T_1$ input}
    \end{subfigure}
     \hspace{0.02\linewidth}
    \begin{subfigure}[b]{0.3\linewidth}
        \includegraphics[width=\linewidth]{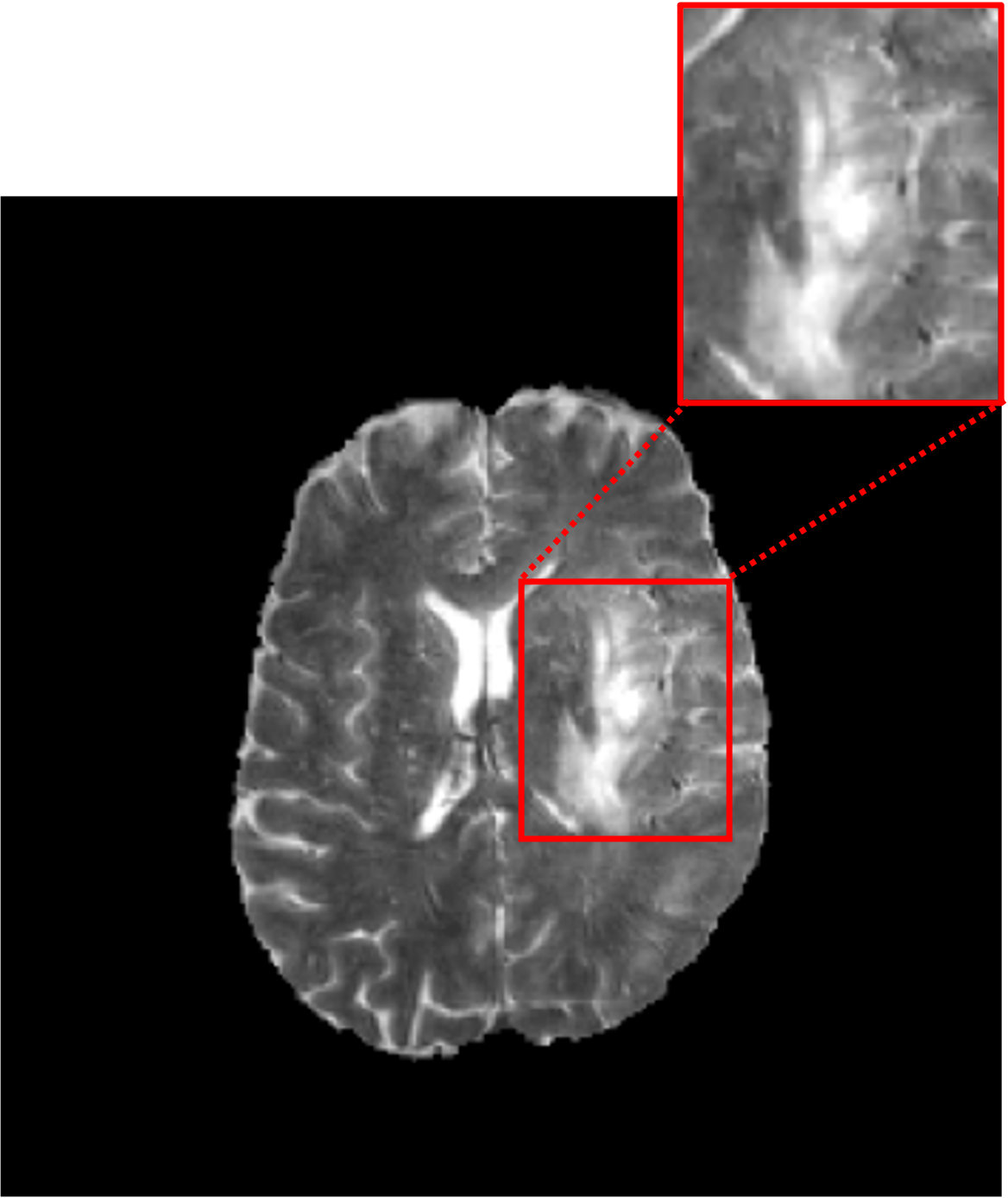}
        \caption{$T_2$ reference}
    \end{subfigure}
    \hspace{0.02\linewidth}
    \begin{subfigure}[b]{0.3\linewidth}
        \includegraphics[width=\linewidth]{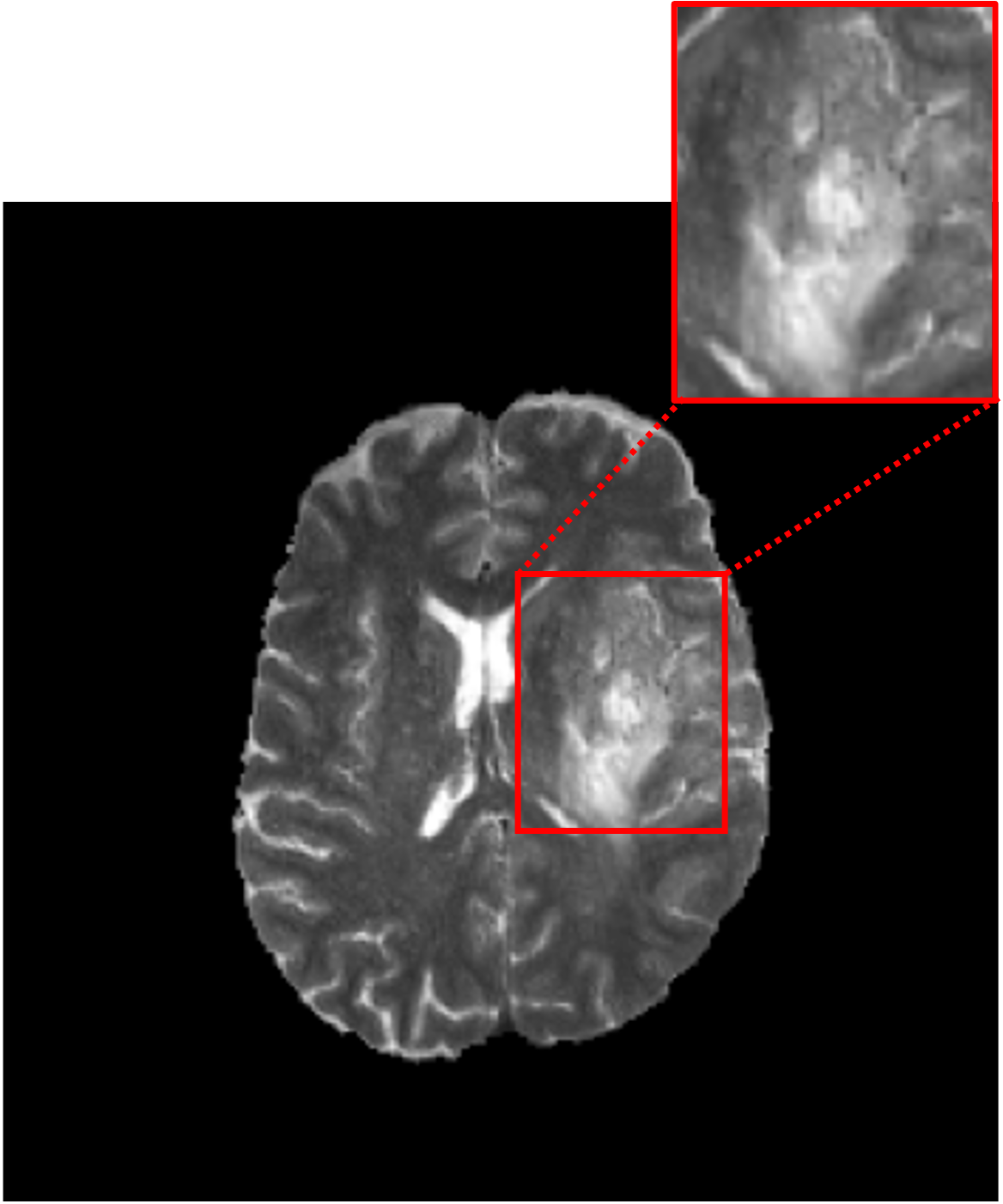}
        \caption{No TTA}
    \end{subfigure}
    
    \vspace{0.4cm}
    
    \begin{subfigure}[b]{0.3\linewidth}
        \includegraphics[width=\linewidth]{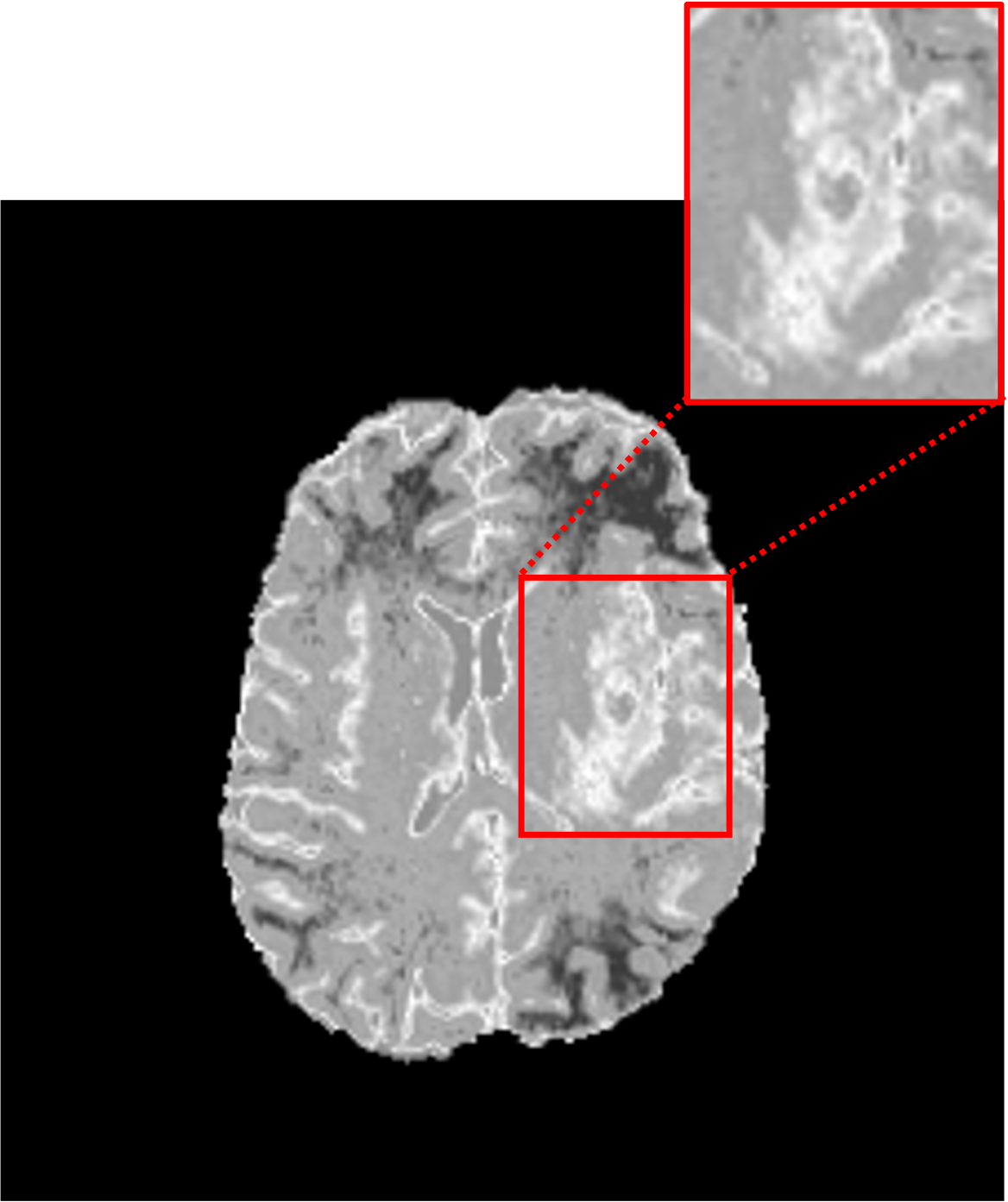}
        \caption{He et al.~\cite{he2021autoencoder}}
    \end{subfigure}
     \hspace{0.02\linewidth}
    \begin{subfigure}[b]{0.3\linewidth}
        \includegraphics[width=\linewidth]{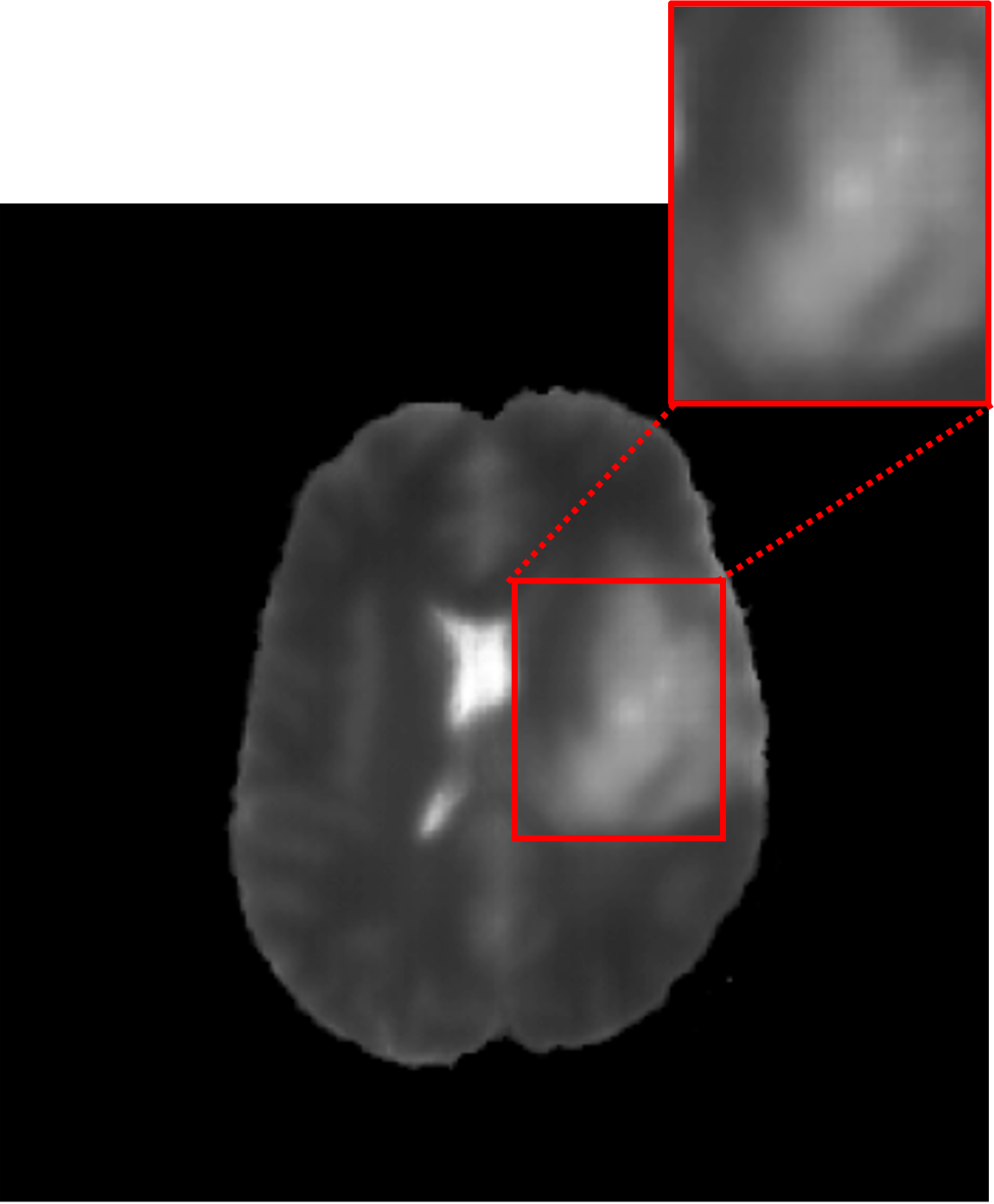}
        \caption{He et al.\textsuperscript{*}}
    \end{subfigure}
     \hspace{0.02\linewidth}
    \begin{subfigure}[b]{0.3\linewidth}
        \includegraphics[width=\linewidth]{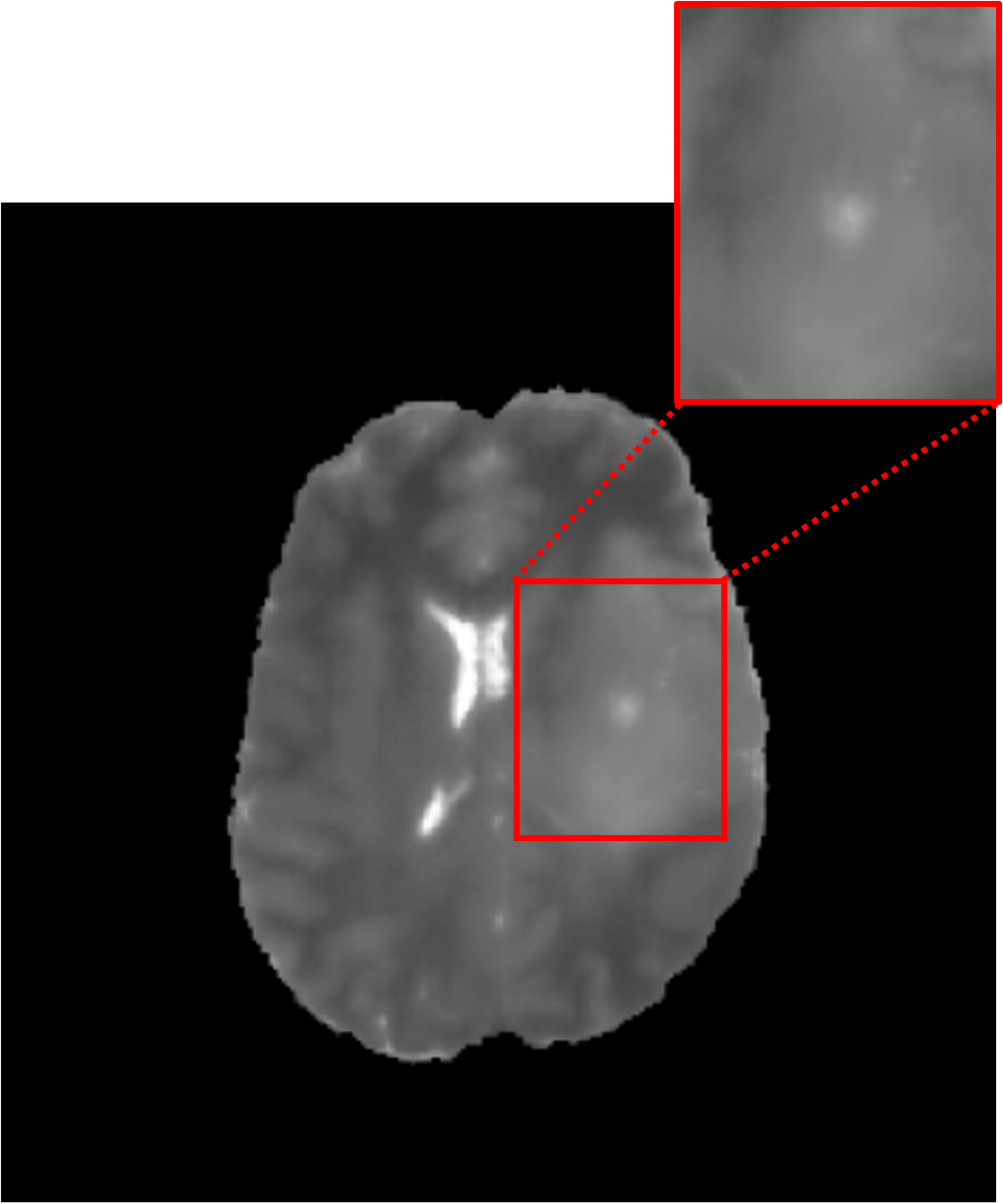}
        \caption{$\mathrm{TTA}_{\mathrm{Grid}}$}
    \end{subfigure}

    \caption{Zoom-in comparison across different TTA strategies for the BraTS 2018 dataset. Zoomed ROIs highlight key regions where TTA does not yield notable improvements in translation quality.}
    \label{fig_BraTS}
\end{figure}

\begin{figure}[ht]
    \centering
    \begin{subfigure}[b]{0.3\linewidth}
        \includegraphics[width=\linewidth]{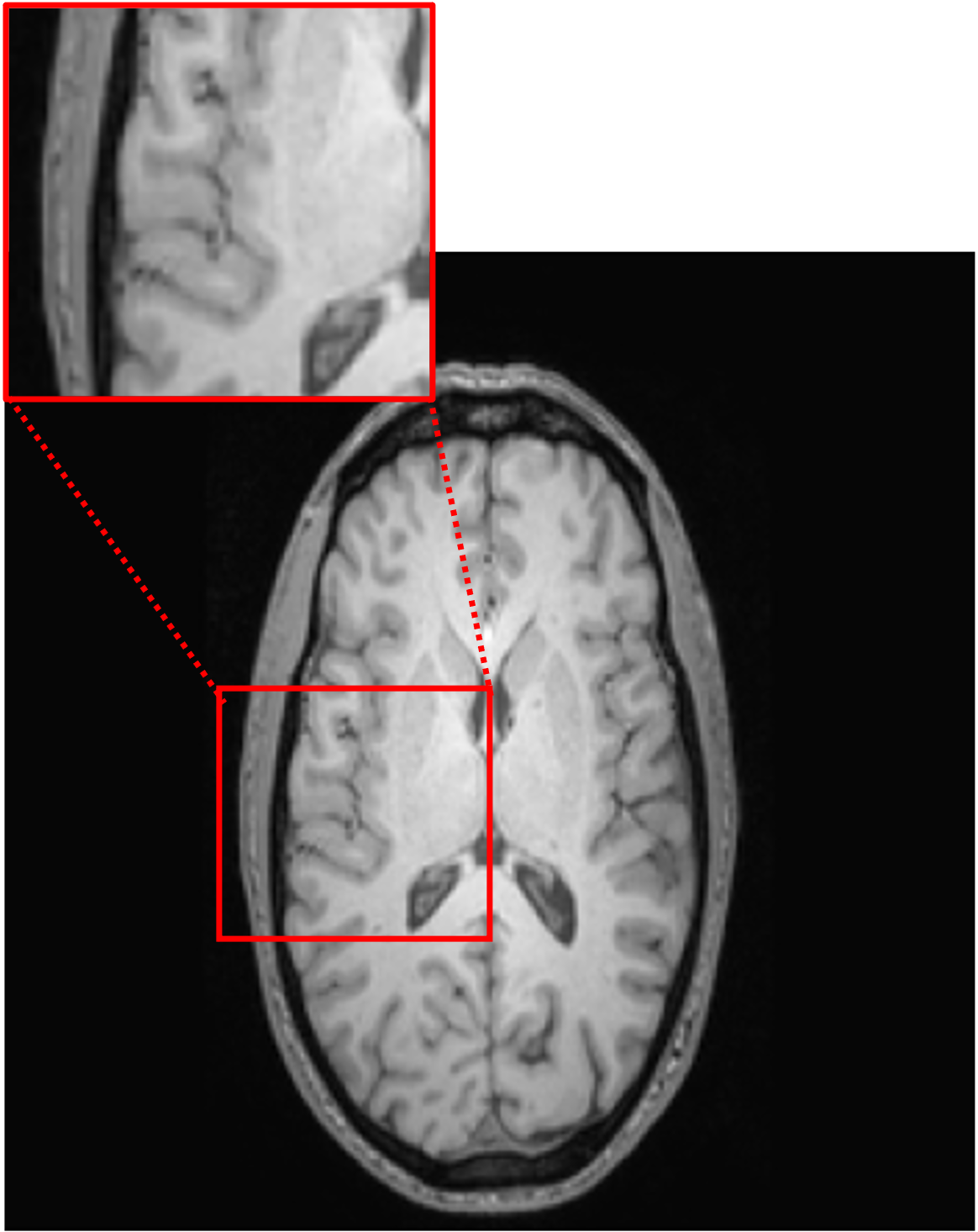}
        \caption{$T_1$ input}
    \end{subfigure}
     \hspace{0.02\linewidth}
    \begin{subfigure}[b]{0.3\linewidth}
        \includegraphics[width=\linewidth]{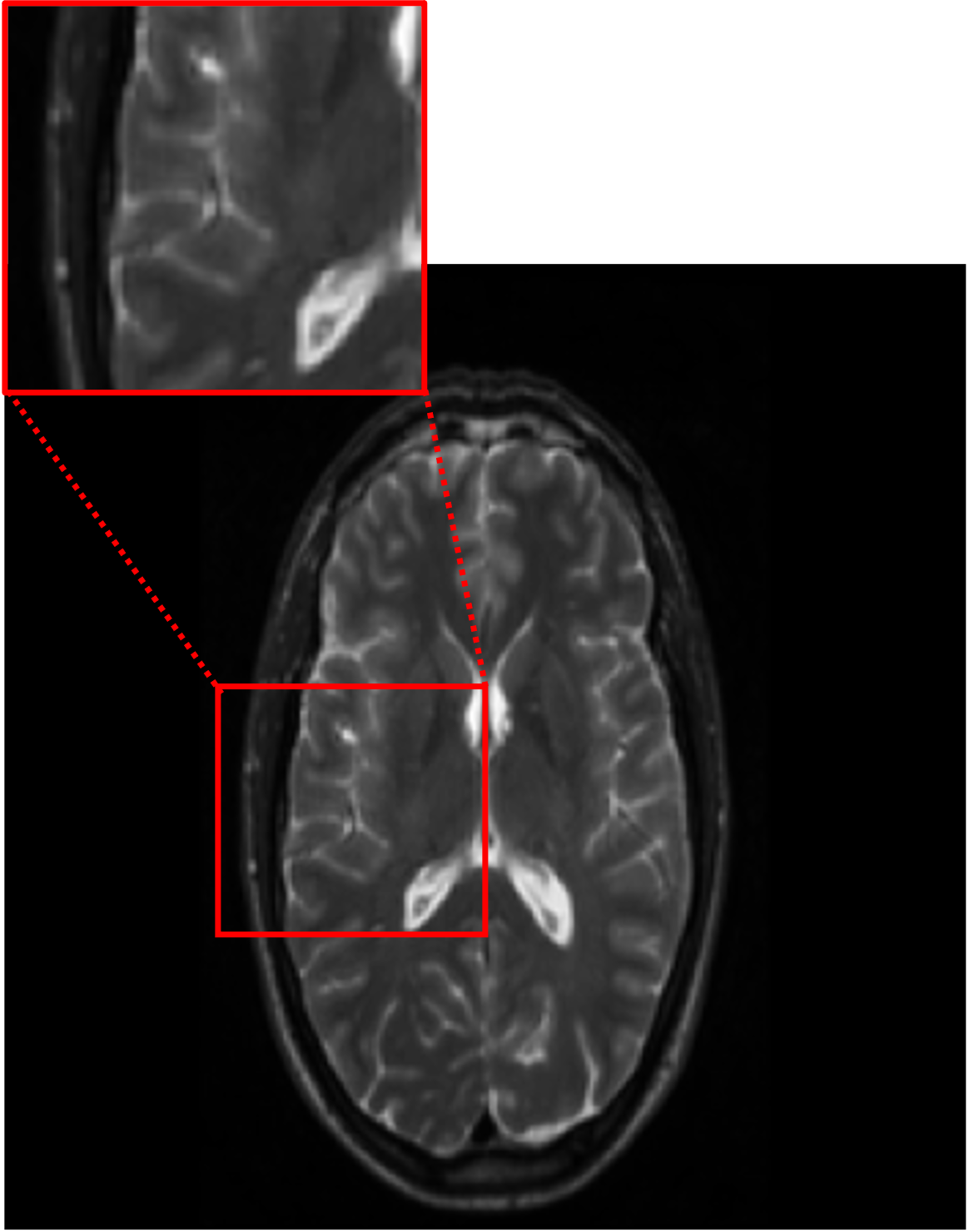}
        \caption{$T_2$ reference}
    \end{subfigure}
    \hspace{0.02\linewidth}
    \begin{subfigure}[b]{0.3\linewidth}
        \includegraphics[width=\linewidth]{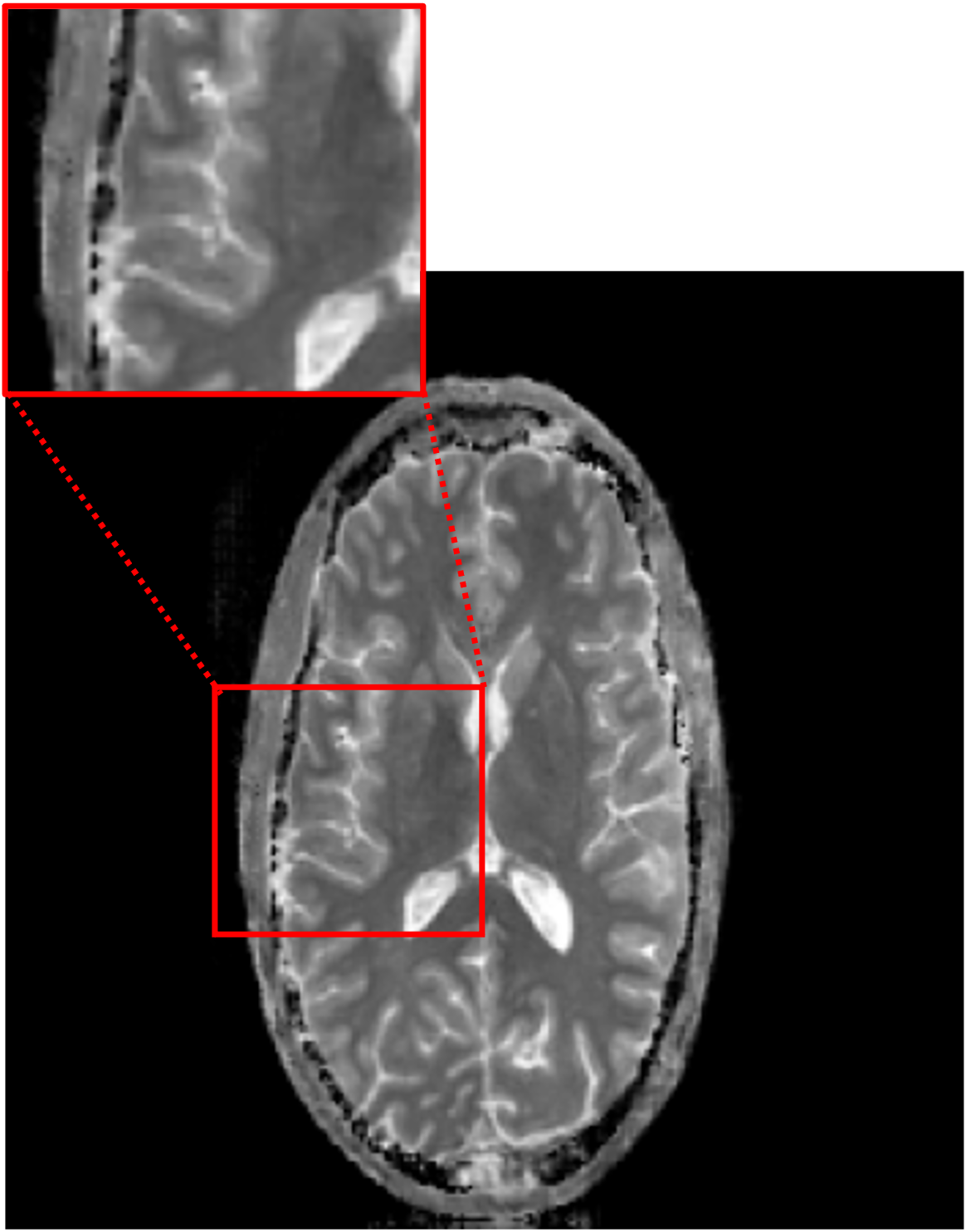}
        \caption{No TTA}
    \end{subfigure}
    
    \vspace{0.4cm}
    
    \begin{subfigure}[b]{0.3\linewidth}
        \includegraphics[width=\linewidth]{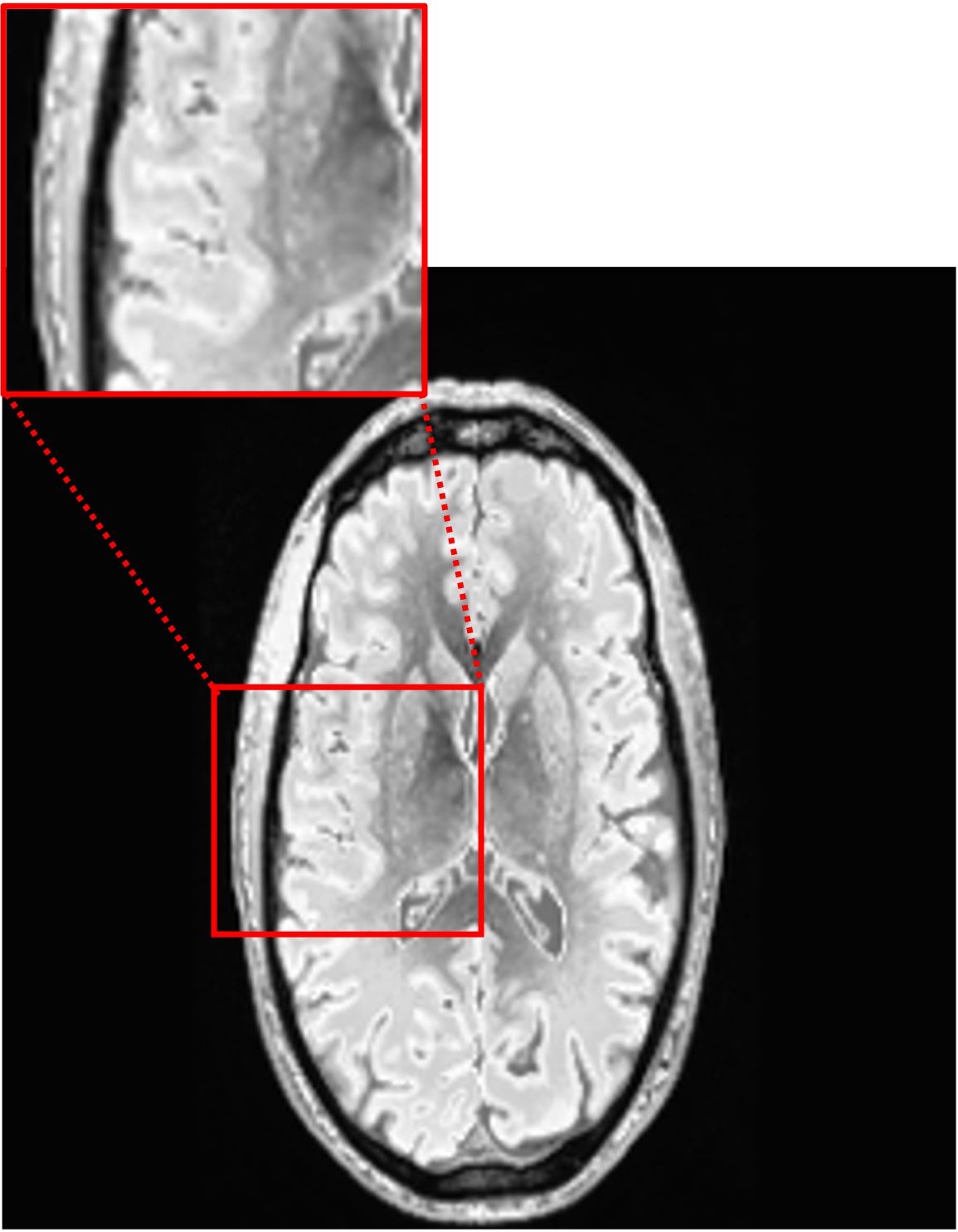}
        \caption{He et al.~\cite{he2021autoencoder}}
    \end{subfigure}
     \hspace{0.02\linewidth}
    \begin{subfigure}[b]{0.3\linewidth}
        \includegraphics[width=\linewidth]{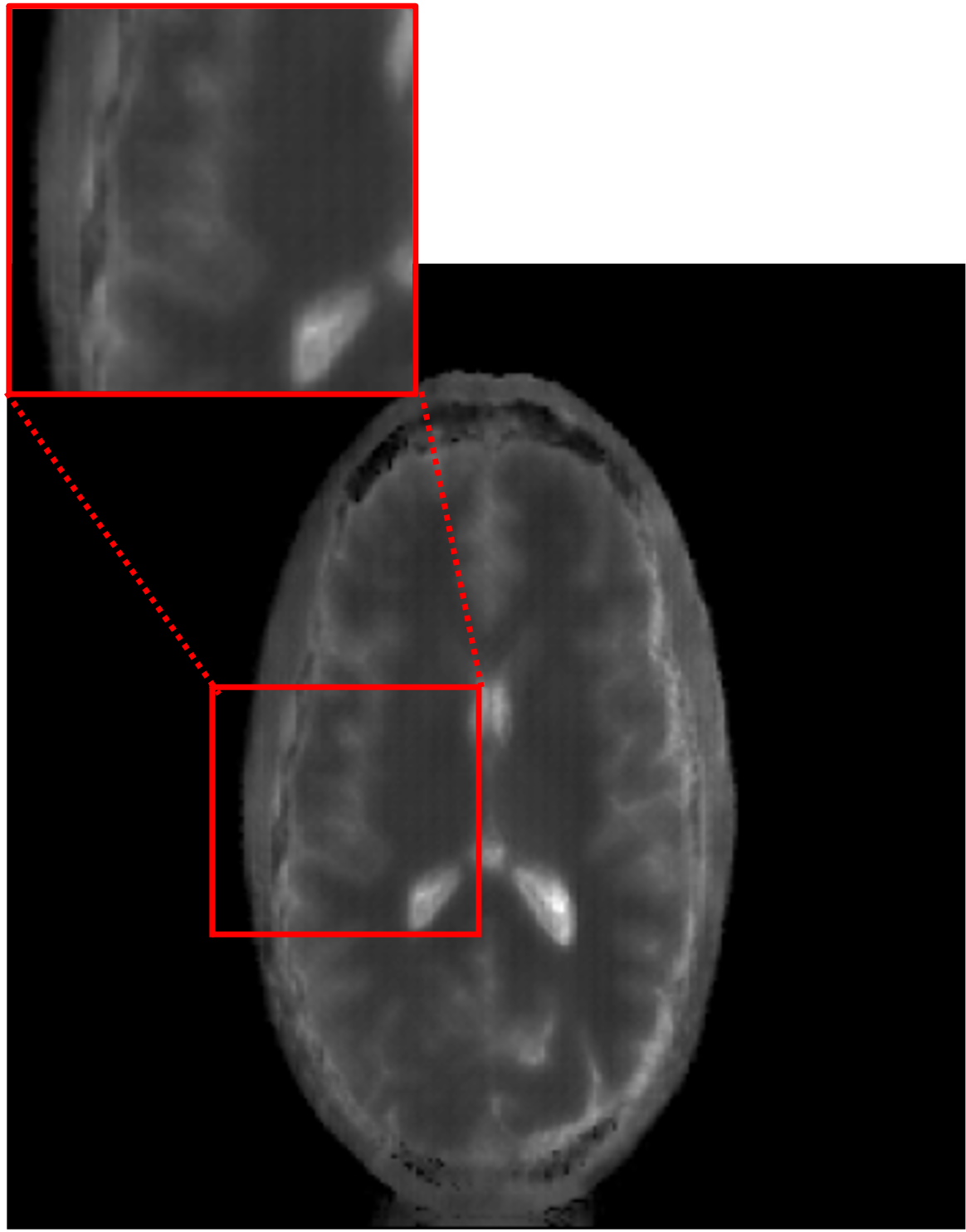}
        \caption{He et al.\textsuperscript{*}}
    \end{subfigure}
     \hspace{0.02\linewidth}
    \begin{subfigure}[b]{0.3\linewidth}
        \includegraphics[width=\linewidth]{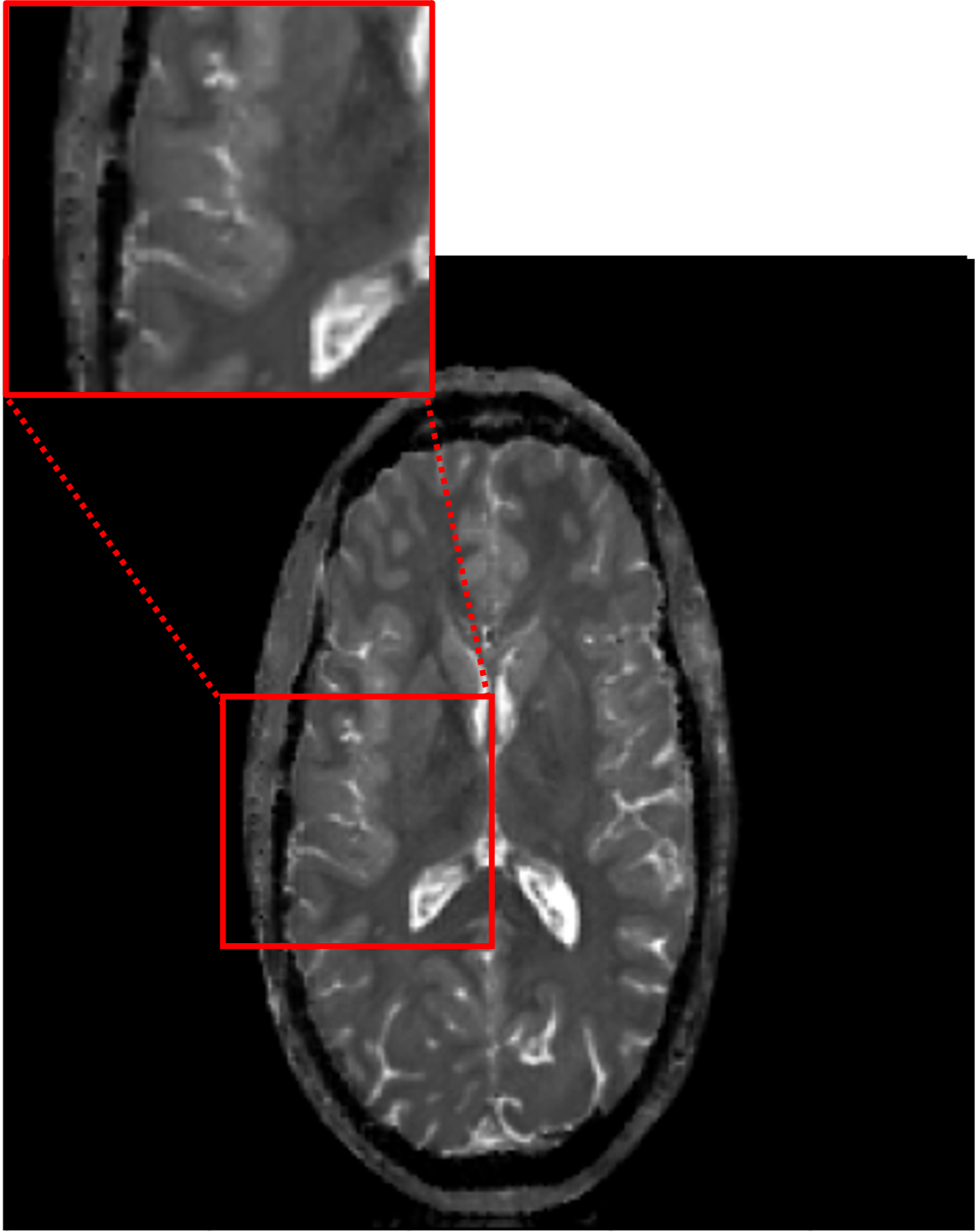}
        \caption{$\mathrm{TTA}_{\mathrm{Grid}}$}
    \end{subfigure}

    \caption{Zoom-in comparison across different TTA strategies for IXI dataset.
        Zoomed-in regions of interest (ROIs) highlight areas where TTA demonstrates notable improvements in translation quality.}
    \label{fig_IXI}
\end{figure}

Figures~\ref{fig_LDCT}, \ref{fig_BraTS}, and \ref{fig_IXI} show visual comparisons for LDCT denoising and $T_1$-to-$T_2$ MRI translation tasks across different experimental configurations.
Each figure follows the same structure: for each dataset, we display the input image, the ground truth reference, and the generated outputs produced by the competitors (No TTA, He et al.~\cite{he2021autoencoder}, and He et al.*) alongside our proposed approach, $\mathrm{TTA}_{\mathrm{Grid}}$. 
Each panel includes a red zoom-in box highlighting a region of interest (ROI) selected to emphasize structural or textural features most affected by the adaptation strategy.

\FloatBarrier

\FloatBarrier

\FloatBarrier

Turning our attention to Figure~\ref{fig_LDCT},  
the denoised image using the task model without adaptation (No TTA, Figure~\ref{fig_LDCT} (c)) exhibits residual noise and structural inconsistencies relative to the high-dose reference (Figure~\ref{fig_LDCT}(b)). 
He et al.~\cite{he2021autoencoder} (Figure~\ref{fig_LDCT} (d)) fails to address these artifacts, while He et al.\textsuperscript{*} (Figure~\ref{fig_LDCT} (e)) yields modest improvements, though it applies a fixed adaptation uniformly across all samples. 
In contrast, $\mathrm{TTA}_{\mathrm{Grid}}$(Figure~\ref{fig_LDCT} (f)) produces improved anatomical fidelity, demonstrating the benefit of sample-specific TTA. 
These results further stress the importance of tailoring TTA to individual test cases rather than relying on static, one-size-fits-all TTA strategies.

Figure~\ref{fig_BraTS} shows results for the $T_1$-to-$T_2$ MRI translation task on the BraTS 2018 dataset. 
In this ID setting, the task model without adaptation (No TTA, Figure~\ref{fig_BraTS} (c)) already generates visually consistent outputs. 
None of the TTA methods, neither the competitors nor our proposed approach, provide noticeable improvements and, in some cases, even degrade image quality.
This reinforces the observation that when dealing with ID test samples, adaptation may be unnecessary and can potentially be detrimental.
In contrast, turning to Figure~\ref{fig_IXI} shows results for the same translation task on the IXI dataset.
Here, the task model model without adaptation (No TTA, Figure~\ref{fig_IXI} (c)) shows evident structural distortions and reduced contrast.
He et al.~\cite{he2021autoencoder} (Figure~\ref{fig_IXI} panel d) fails to correct these artifacts, while He et al.\textsuperscript{*} (Figure~\ref{fig_IXI} (e)) achieves only modest improvements.
In contrast, $\mathrm{TTA}_{\mathrm{Grid}}$ (Figure~\ref{fig_IXI} (f)) delivers the highest visual quality, recovering sharper anatomical boundaries and minimizing artifacts.
Our thorough analysis confirms the core intuition behind sample-aware TTA: while adaptation offers limited benefit ID settings, it becomes essential for improving generalization under significant distribution shifts, provided it is applied selectively and tailored to individual test samples.

\subsection{Alternative Search Strategies for Sample-Aware TTA}
\label{Sec:Alternative Search Strategies}
To address the computational overhead of $\mathrm{TTA}_{\mathrm{Grid}}$, we investigate alternative search strategies for dynamic adaptor selection, each differing in how they explore the configuration space $\Omega$.
As detailed in Table~\ref{tab:search_strategies}, we evaluate six representative search strategies, spanning heuristic and probabilistic approaches, including random searches, named as $\mathrm{TTA}_{\mathrm{R10}}$ and $\mathrm{TTA}_{\mathrm{R50}}$, which sample 10 and 50 configurations, respectively; forward selection $\mathrm{TTA}_{\mathrm{FS}}$, backward elimination $\mathrm{TTA}_{\mathrm{BE}}$, and bayesian optimization $\mathrm{TTA}_{\mathrm{BA}}$.
\begin{table}[ht]
\centering
\small
\begin{tabular}{lm{10cm}}
\toprule
 \textbf{Experiment} & \textbf{Description} \\
\midrule
$\mathrm{TTA}_{\mathrm{R10}}$ & Sample-aware with random search over 10 sampled configurations. \\
$\mathrm{TTA}_{\mathrm{R50}}$ & Sample-aware TTA with random search over 50 sampled configurations. \\
$\mathrm{TTA}_{\mathrm{FS}}$ & Sample-aware TTA with forward selection search. \\
$\mathrm{TTA}_{\mathrm{BE}}$ & Sample-aware TTA with backward elimination search. \\
$\mathrm{TTA}_{\mathrm{BA}}$ & Sample-aware TTA with Bayesian search.  \\
\bottomrule
\end{tabular}
\caption{Summary of alternative search strategies used for sample-aware TTA.}
\label{tab:search_strategies}
\end{table}

%\subsubsection{LDCT denoising}
\label{sec:denoising}
Table~\ref{tab:search_strategies_} reports the quantitative metrics in the same format as Table~\ref{tab:tasks}. 
Results on BraTS 2018 are omitted, as prior analysis showed that TTA is ineffective in this setting, making further exploration unnecessary.
The table is divided into two sections according to the translation task: LDCT denoising, and $T_1$ to $T_2$ translation on the IXI dataset.
We report the results obtained by each method on both the entire test set $\bm{A}$ and the subset of OOD samples $\bm{B} \subset \bm{A}$.
In the LDCT denoising task, $\mathrm{TTA}_{\mathrm{Grid}}$ demonstrates strong overall performance but is not consistently statistically superior to other search strategies, particularly $\mathrm{TTA}_{\mathrm{R50}}$, as detailed in~\ref{Appendix: statistical Analysis}.
In contrast, greedy algorithms such as $\mathrm{TTA}_{\mathrm{FS}}$ and $\mathrm{TTA}_{\mathrm{BE}}$ exhibit greater performance variability, likely due to their tendency to get stuck in local minima within the search space.
Bayesian optimization, $\mathrm{TTA}_{\mathrm{BA}}$, performs reasonably well but does not outperform simpler alternatives, possibly due to the limited number of trials and the need for careful hyperparameter tuning, which itself introduces additional complexity.
Turning to the $T_1$-to-$T_2$ MRI translation task, we observe a similar pattern: while $\mathrm{TTA}_{\mathrm{Grid}}$ maintains strong performance, its advantage over lighter strategies, such as $\mathrm{TTA}_{\mathrm{R50}}$ and even $\mathrm{TTA}_{\mathrm{R10}}$, is not always statistically significant, as detailed in \ref{Appendix: statistical Analysis}.
These findings underscore the effectiveness of random search methods, even with a small number of evaluations, in achieving competitive performance with significant reduction in computational cost.

\begin{table*}[!h]
\centering
%\begin{adjustbox}{center}
%\resizebox{16.5cm}{!}{
\resizebox{1\textwidth}{!}{
\begin{tabular}{l|l|cc|cc|cc}
\toprule
\textbf{Task} & \textbf{Experiment} & \multicolumn{2}{c|}{\textbf{SSIM ↑}} 
& \multicolumn{2}{c|}{\textbf{MAE ↓}} 
& \multicolumn{2}{c}{\textbf{PSNR ↑}} \\
\toprule
 & \textbf{}  
& \textbf{$\bm A$} & $\bm{B \subset A}$ 
& \textbf{$\bm A$} & $\bm{B \subset A}$ 
& \textbf{$\bm A$} & $\bm{B \subset A}$ \\
\toprule
\multirow{6}{4em}{LDCT \\ Denoising} &  $\mathrm{TTA}_{\mathrm{Grid}}$ & \cellcolor{lightgreen} .699 \textsuperscript{$\pm$ .203} & \cellcolor{lightgreen} .769 \textsuperscript{$\pm$ .289} & \cellcolor{lightgreen} .060 \textsuperscript{$\pm$ .045} & \cellcolor{lightgreen} .063 \textsuperscript{$\pm$ .053} & \cellcolor{lightgreen} 28.464 \textsuperscript{$\pm$ 5.492} & \cellcolor{lightgreen} 29.204 \textsuperscript{$\pm$ 6.137} \\
& 
$\mathrm{TTA}_{\mathrm{R10}}$*  & .699 \textsuperscript{$\pm$ .202} & .761 \textsuperscript{$\pm$ .286} & .061 \textsuperscript{$\pm$ .045} & .068 \textsuperscript{$\pm$ .052} & 28.411 \textsuperscript{$\pm$ 5.471} & 28.139 \textsuperscript{$\pm$ 5.779} \\
  & $\mathrm{TTA}_{\mathrm{R50}}$*  &\cellcolor{lightblue}  .699 \textsuperscript{$\pm$ .202} &\cellcolor{lightblue}  .764 \textsuperscript{$\pm$ .286} &\cellcolor{lightblue} .061 \textsuperscript{$\pm$ .045} &\cellcolor{lightblue}  .066 \textsuperscript{$\pm$ .052} &\cellcolor{lightblue}  28.435\textsuperscript{$\pm$ 5.477} &\cellcolor{lightblue}  28.625 \textsuperscript{$\pm$ 5.899} \\
  & $\mathrm{TTA}_{\mathrm{FS}}$  & .698 \textsuperscript{$\pm$ .203} & .745 \textsuperscript{$\pm$ .294} & .062 \textsuperscript{$\pm$ .048} & .084 \textsuperscript{$\pm$ .088} & 28.342 \textsuperscript{$\pm$ 5.534} & 26.737 \textsuperscript{$\pm$6.705} \\
  & $\mathrm{TTA}_{\mathrm{BE}}$ & .699 \textsuperscript{$\pm$ .202} & .759 \textsuperscript{$\pm$ .284} & .061 \textsuperscript{$\pm$ .045} & .071 \textsuperscript{$\pm$ .058} & 28.404 \textsuperscript{$\pm$ 5.488} & 27.980 \textsuperscript{$\pm$ 6.095} \\
  & $\mathrm{TTA}_{\mathrm{BA}}$ & .697 \textsuperscript{$\pm$.201} & .731 \textsuperscript{$\pm$ .275}  & .062 \textsuperscript{$\pm$ .047} & .087 \textsuperscript{$\pm$ .079} & 28.293 \textsuperscript{$\pm$ 5.484} & 25.745 \textsuperscript{$\pm$ 5.450}  \\ 
\midrule\midrule
\multirow{6}{4em}{MRI $T_1$-$T_2$ \\(IXI)} &  $\mathrm{TTA}_{\mathrm{Grid}}$ & \cellcolor{lightgreen} .739 \textsuperscript{$\pm$ .068} & \cellcolor{lightgreen}.771 \textsuperscript{$\pm$ .060} & .093 \textsuperscript{$\pm$ .022} & .088 \textsuperscript{$\pm$ .024} & \cellcolor{lightgreen} 20.042 \textsuperscript{$\pm$ 1.968} & \cellcolor{lightgreen} 20.317 \textsuperscript{$\pm$ 2.275} \\
& $\mathrm{TTA}_{\mathrm{R10}}$*  &\cellcolor{lightblue} .729 \textsuperscript{$\pm$ .060} &\cellcolor{lightblue} .760 \textsuperscript{$\pm$ .048} & \cellcolor{lightgreen}.088 \textsuperscript{$\pm$ .022} & \cellcolor{lightgreen}.081 \textsuperscript{$\pm$ .022} & 19.303 \textsuperscript{$\pm$ 2.328} & 19.447 \textsuperscript{$\pm$ 1.946}\\
& $\mathrm{TTA}_{\mathrm{R50}}$*  & .727 \textsuperscript{$\pm$ .063} & .758 \textsuperscript{$\pm$ .056} &\cellcolor{lightblue} .089 \textsuperscript{$\pm$ .021} &\cellcolor{lightblue} .082 \textsuperscript{$\pm$ .022} &\cellcolor{lightblue} 19.676 \textsuperscript{$\pm$ 1.899} &\cellcolor{lightblue} 19.678 \textsuperscript{$\pm$ 2.275} \\
& $\mathrm{TTA}_{\mathrm{FS}}$  & .716 \textsuperscript{$\pm$ .059} & .740 \textsuperscript{$\pm$ .057} & .099 \textsuperscript{$\pm$ .021} & .100 \textsuperscript{$\pm$ .024} & 18.850 \textsuperscript{$\pm$ 1.834} & 18.325 \textsuperscript{$\pm$ 2.019} \\
& $\mathrm{TTA}_{\mathrm{BE}}$ & .728 \textsuperscript{$\pm$ .060} & .759 \textsuperscript{$\pm$ .050} & .090 \textsuperscript{$\pm$ .020} & .083 \textsuperscript{$\pm$ .020} & 19.435 \textsuperscript{$\pm$ 2.022} & 19.284 \textsuperscript{$\pm$ 2.430} \\
& $\mathrm{TTA}_{\mathrm{BA}}$ & .721 \textsuperscript{$\pm$ .054} & .747 \textsuperscript{$\pm$ .045} & .094 \textsuperscript{$\pm$.021} & .091 \textsuperscript{$\pm$.024} & 18.925 \textsuperscript{$\pm$1.853} & 18.448 \textsuperscript{$\pm$ 2.077} \\ 
\bottomrule
\end{tabular}
}
%\end{adjustbox}
\caption{Quantitative comparison of different search strategies for sample-aware TTA. 
Metrics are reported for the entire test set $\bm{A}$ and for the subset $\bm{B} \subset \bm{A}$. 
The best results for each task are highlighted in green, while the second-best are marked in blue.
An asterisk (*) denotes that results were averaged over three runs to reduce variance and improve stability.
The dagger symbol ($^\dagger$) denotes results from configurations that do not support OOD sample identification but are evaluated on the subset $\bm{B} \subset \bm{A}$, identified by our approach, to illustrate how these configurations would perform on the detected OOD samples.
} 
%The best results are marked in bold and their statistical significance is assessed with Wilcoxon signed rank test ($p < 0.05$).}
\label{tab:search_strategies_}
\end{table*}
\FloatBarrier

\subsection{Computational analysis}
\label{sec: Computational analysis}
Turning our attention to the computational cost, Table~\ref{tab:inference_time} shows the average inference time and complexity $\mathcal{O}(\cdot)$ computed on a single A100 across all tasks.
The analysis accounts for both the adaptor update steps and the configuration selection process involved in each TTA configuration, along with their corresponding computational complexity.
As expected, all alternative search strategies introduce additional computational overhead compared to the baseline.
Among them, $\mathrm{TTA}_{\mathrm{Grid}}$ is by far the most computationally demanding, requiring up to 130 seconds (s) per sample. 
This is due to its exhaustive exploration of the configuration space $\Omega$, whose complexity grows exponentially with the number of intermediate layers $n{-}2$, resulting in $\mathcal{O}(2^{n-2})$ operations for each test samples requiring adaptation.
To mitigate this cost, we evaluate alternative strategies that reduce the number of configurations explored while maintaining dynamic and sample-specific adaptation. 
Random searches, $\mathrm{TTA}_{\mathrm{R10}}$ and $\mathrm{TTA}_{\mathrm{R50}}$, with $ N_{config}$ equal to 10 and 50, respectively, limit the search to 10 or 50 configurations, respectively, achieving inference times of 12 s and 55 s.
As expected, their complexity scales linearly with the number of sampled configurations, i.e., $\mathcal{O}(N_{config})$, offering a trade-off between speed and performance.
$\mathrm{TTA}_{\mathrm{FS}}$ and $\mathrm{TTA}_{\mathrm{BE}}$ achieve a more favorable balance, with inference times around 25 s and 21 s. 
Their quadratic complexity $\mathcal{O}((n-2)^2)$ reflects the structured, iterative nature of the search over $n{-}2$ intermediate layers. 
Finally, $\mathrm{TTA}_{\mathrm{BA}}$ offers a principled compromise, with a fixed budget of $N_{config}=20$ trials guided by a probabilistic surrogate model. 
While its complexity remains linear in the number of evaluations, the targeted search effectively avoids unnecessary configurations, resulting in inference times of approximately 20 s.
While $\mathrm{TTA}_{\mathrm{Grid}}$ remains the most exhaustive and frequently top-performing strategy, our statistical analysis reveals that its performance is not always consistently or significantly superior to lighter alternatives.
In particular, approaches such as $\mathrm{TTA}_{\mathrm{R50}}$ often achieve comparable results without incurring the high computational cost of a grid search.
This suggests that, in practice, more efficient search strategies can provide a favorable trade-off between adaptation quality and inference time, offering competitive performance with significantly lower resource demands.

\begin{table}[h!]
\centering
\begin{tabular}{l l c c}
\toprule
\multicolumn{2}{c}{\textbf{TTA}} & \textbf{Inference Time (s/sample)} & $\bm{{\mathcal{O}}(\cdot)}$ \\
\midrule
& $\mathrm{TTA}_{\mathrm{Grid}}$ & 130  & $\mathcal{O}(2^{n-2})$\\
  & $\mathrm{TTA}_{\mathrm{R10}}$  & 12 & $\mathcal{O}(N_{config})$ \\
  & $\mathrm{TTA}_{\mathrm{R50}}$  & 55 & $\mathcal{O}(N_{config})$  \\
  & $\mathrm{TTA}_{\mathrm{FS}}$   & 25 & $\mathcal{O}((n-2)^2)$  \\
  & $\mathrm{TTA}_{\mathrm{BE}}$   & 21 & $\mathcal{O}((n-2)^2)$ \\
  & $\mathrm{TTA}_{\mathrm{BA}}$  & 20 & $\mathcal{O}(N_{config})$  \\
\bottomrule
\end{tabular}
\caption{Average inference time per sample (in seconds), measured across all tasks. 
The reported times include both adaptor update and the configuration selection search. 
The last column reports the theoretical complexity $\mathcal{O}(\cdot)$ per sample,  where $n$ denotes the total number of layers in the task model, $n{-}2$ denotes the number of intermediate layers and $N_{config}$ the number of explored configurations}

\label{tab:inference_time}
\end{table}
\FloatBarrier

\section{Conclusions}
\label{sec:conclusion}
In this paper, we introduce an approach for sample-aware TTA in medical image-to-image translation.
The core contributions of this work are twofold: (1) a reconstruction module that quantifies reconstruction errors across all feature levels of a pretrained task model $\mathcal{T}$, serving as a proxy for domain shift; and (2) a dynamic adaptation block that applies feature-level transformations at multiple stages of $\mathcal{T}$, guided by the reconstruction errors identified by the reconstruction module.
Through extensive evaluation on LDCT denoising and $T_1$-to-$T_2$ MRI translation tasks, we demonstrate that TTA is most effective when applied selectively. While indiscriminate adaptation can degrade performance on ID samples, our method delivers substantial improvements on OOD cases by tailoring the adaptation on a per-sample basis.
We show that dynamically selecting the optimal adaptation configuration on a per-sample basis outperforms static, one-size-fits-all approaches in both quantitative metrics and computational efficiency.
These findings highlight the importance of balancing adaptation quality with runtime constraints, a critical consideration for real-world deployment.

\paragraph{Limitations and Future Work}
Despite these promising results, our approach has some limitations.
First, OOD detection relies on a fixed percentile threshold applied to reconstruction error, which may not generalize optimally across tasks or domains.
Second, although we mitigate the overhead of configuration search, the process still introduces latency that may limit its use in real-time or resource-constrained settings.
Third, our current implementation operates on 2D slices, consistent with most prior work in medical image-to-image translation.
This choice reflects not only computational considerations, but also our primary goal: to introduce and systematically evaluate a novel TTA strategy, rather than to optimize translation performance. 
Working in 2D enables controlled benchmarking of sample-specific adaptation mechanisms. 
However, this formulation does not capture volumetric consistency, which remains essential for many clinical applications.
As a first direction for future work, we aim to improve OOD detection by exploring more advanced selection strategies, such as learned thresholds, outlier detection techniques, or uncertainty-based scoring.
Additionally, we will investigate adaptive mechanisms for more efficient and generalizable adaptor selection across diverse generative tasks and data modalities.
We also plan to extend our framework to a broader range of generative architectures such as diffusion models or vision transformers, evaluating its model-agnostic capabilities and its robustness in real-time and multi-modal deployment scenarios.
Lastly, we plan to extend our framework to 3D architectures such as volumetric GANs or diffusion models, leveraging the modularity of our adaptation design to support fully volumetric adaptation.

\section{Acknowledgment}
Irene Iele is a Ph.D. student enrolled in the National Ph.D. in Artificial Intelligence, XL cycle, course on Health and Life Sciences, organized by Università Campus Bio-Medico di Roma.
This work was partially funded by: i) Kempe Foundation project JCSMK24-0094; ii) Cancerforskningsfonden Norrland project MP23-1122; iii) PNRR-MCNT2-2023-12377755 LUMINATE.
The computations of this work were enabled by resources provided by the Swedish National Infrastructure for Computing (SNIC), partially funded by the Swedish Research Council through grant agreement no. 2018-05973.  
Generative AI tools were used to assist in grammar correction and text style refinement throughout the paper.

%\bibliographystyle{elsarticle-num-names} 
%\bibliography{biblio}

\newpage
\appendix
\section{Preliminaries}
\label{appendix:Preliminaries}
\subsection{CycleGAN}
CycleGAN \cite{zhu2017unpaired}, introduced in 2017, made significant changes to the traditional GANs' architecture \cite{goodfellow2020generative}, enabling bidirectional image-to-image translation. 
Unlike standard GANs, which focus on unidirectional translation, CycleGAN revolutionizes this paradigm, allowing both the translation from the source domain to the target domain and the reverse translation from synthetic target images back to the source domain.
This dual transformation enhances model stability and improves the accuracy of generated images, especially in unpaired settings, where paired training data is unavailable.
\begin{figure}[ht]
\centering
\includegraphics[width=\textwidth]{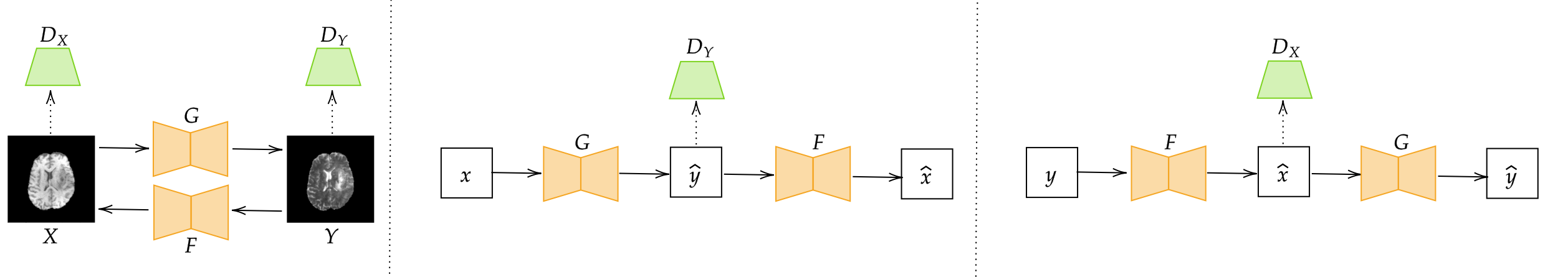}
\caption{CycleGAN Architecture.}
\label{fig:Cyclegan}
\end{figure} 
The CycleGAN framework, as shown in figure~\ref{fig:Cyclegan} includes two mapping functions, known as generators: $G:X \to Y $ and $F: Y \to X $, where $X$ is the source image domain while $Y$ is the target image domain. 
It also includes two discriminators, $D_X$ and $D_Y$, which evaluate the authenticity of images translated between these domains. 
In particular, the role of the discriminator $D_Y$ is to differentiate between samples generated by $G(x)$ and authentic samples from domain $Y$, while $D_X$ distinguishes between generated samples $F(y)$ and the real samples of the $X$ domain.
CycleGAN is composed of two loops: the consistency of the forward cycle and the consistency of the backward cycle.
The first loop is a mapping from the source image domain $X$ to the target image domain $Y$ and then a mapping from the target image domain back to the source image domain; therefore, the image translation process must ensure that any image $x$ belonging to $X$, when translated into domain $Y$, and then returned to domain $X$, remains approximately the same original image $x$.
\begin{equation}
     x \to G(x) = \hat{y} \to F(\hat{y}) = \hat{x}.
\end{equation} 
Similarly, the second loop is a mapping from the target image domain to the source image domain and then a mapping from the source image domain back to the target image domain. So, for any image $y$ from domain $Y$, the process should ensure that after being translated to domain $X$, and then back to domain $Y$, it remains almost identical to the original image $y$.
\begin{equation} y \to F(y) = \hat{x} \to G(\hat{x}) = \hat{y}. \end{equation}
The CycleGAN optimization function is formulated as a minimax problem, in which the generators and discriminators are trained simultaneously. 
The goal is to find the optimal parameters for both the generators and discriminators that minimize the ability of the discriminators to distinguish between generated and real samples. The minimax formulation is expressed as:
\begin{equation}\min_{G, F} \max_{D_X, D_Y} \mathcal{L}_{\text{GAN}}(G, F, D_X, D_Y). \end{equation}
The performance of these networks is therefore highly dependent on how their loss functions are structured, so special attention must be paid to them.
The loss used in CycleGAN is composed of three terms: adversarial loss, cycle consistency loss and identity loss.
The first term is applied to both mapping functions. 
For the mapping function $G : X \to Y$ and its discriminator $D_Y$, we express the objective as: where $G$ tries to generate images $G(x)$ that look similar to images from domain $Y$, while $D_Y$ aims to distinguish between translated samples $G(x)$ and real samples $y$.
\begin{equation}
\mathcal{L}_{\text{GAN}}(G, D_Y, X, Y) = \mathbb{E}_{y \sim p_{\text{data}}(y)} \left[ \log D_Y(y) \right] 
+ \mathbb{E}_{x \sim p_{\text{data}}(x)} \left[ \log \left( 1 - D_Y(G(x)) \right) \right].
\label{eq:gan_1}
\end{equation}
For the mapping function $F : Y \to X$ and its discriminator $D_X$ as well we have:
\begin{equation}
\mathcal{L}_{\text{GAN}}(F, D_x, Y, X) = \mathbb{E}_{x \sim p_{\text{data}}(x)} \left[ \log D_X(x) \right] 
+ \mathbb{E}_{y \sim p_{\text{data}}(y)} \left[ \log \left( 1 - D_X(F(y)) \right) \right].
\label{eq:gan_2}
\end{equation}
Where $\mathbb{E}_{y \sim p_{\text{data}}(y)}$,  $\mathbb{E}_{x \sim p_{\text{data}}(x)}$ represent, respectively, the expectation with respect to the distribution of real images in the $Y$ and $X$ domains. $D_Y(y)$ is the probability that image $y$ is deemed authentic in domain $Y$ by discriminator $D_Y$; $D_Y(G(x))$ represents the probability that the discriminator $D_Y$ classifies the generated images $G(x)$ as authentic within domain $Y$.
\\On the other hand, $D_X(x)$ is the probability that image $x$ is deemed authentic in domain $X$ by discriminator $D_X$ and $D_X(F(y))$ represents the probability that the discriminator $D_X$ classifies the generated images $F(y)$ as authentic within domain $X$.
${\mathcal{L}}_{\text{adv}}$ is the sum of~\ref{eq:gan_1} and~\ref{eq:gan_2} expressions.
\\In CycleGANs, compared to traditional GANs, a new loss, known as Cycle Consistency Loss, is introduced, which further regularizes the mappings: for each image $x$ of domain $X$, the image translation cycle must be able to bring $x$ back to the original image. This is called forward cycle consistency. Similarly, for each image $y$ from domain $Y,G$ and $F$ should also satisfy backward cycle consistency. \bb
\begin{equation}
\mathcal{L}_{\text{cyc}}(G, F) = \mathbb{E}_{x \sim p_{\text{data}}(x)} \left[ \| F(G(x)) - x \|_1 \right] 
+ \mathbb{E}_{y \sim p_{\text{data}}(y)} \left[ \| G(F(y)) - y \|_1 \right]
\end{equation}
In this case, the loss is computed using the L1 norm.
\\
The third term, the Identity loss, acts as an additional parameter aimed at elevating the overall quality of the generated images. 
This loss has special relevance to avoid unwanted modifications when an image already belongs to the target domain.
The essence of identity loss is to take a sample $x$ from the target domain $X$ and pass it through generator $G$. 
Similarly, a sample y from the target domain $Y$ is passed through generator $F$. 
In this way, both generators learn the identity mapping functions for their respective domains. 
The equation representing the loss of identity is expressed as follows:
\begin{equation}
\mathcal{L}_{\text{identity}}(G, F) = \mathbb{E}_{x \sim p_{\text{data}}(x)} \left[ \| G(x) - x \|_1 \right] 
+ \mathbb{E}_{y \sim p_{\text{data}}(y)} \left[ \|F(y) - y \|_1 \right]
\end{equation}
Also in this case, the loss is computed using the L1 norm.
So we can resume that the CycleGAN loss is:
\begin{equation}
\mathcal{L}_{\text{GAN}}(G, F, D_X, D_Y) ={L}_{\text{adv}}(G, F, D_X, D_Y) + \lambda_1{L}_{\text{cycle}}(G, F) +  \lambda_2{L}_{\text{identity}}(G, F)
\end{equation}
%\red Our task model is only the CycleGan's generator. This Resnet Generator has fifteen depth level and nine resnet blocks. \bb
\subsection{Autoencoder}
Autoencoders \cite{rumelhart1986parallel} are a class of neural networks designed for unsupervised learning, in which the primary goal is to learn an efficient encoding of the input data while allowing accurate reconstruction of the original input.
This process makes them particularly suitable for tasks aimed at dimensionality reduction, feature extraction, and data reconstruction.
An Autoencoder consists of two main modules:
\begin{itemize}
    \item Encoder: This module compresses the input data into a compact latent representation, capturing the most salient and essential features of the data. The encoder reduces the input dimensionality, discarding irrelevant or redundant information.
    \item Decoder: Reconstructs the input data from the compressed latent representation. Its goal is to recreate the original data as faithfully as possible, effectively reversing the transformation applied by the encoder.
\end{itemize}

Autoencoder training is controlled by a reconstruction loss function, which measures the difference between the input data and its reconstruction. The choice of loss function significantly affects the model's ability to learn relevant features. Commonly used loss functions include Mean Squared Error (MSE), Mean Absolute Error (MAE) or Structural Similarity Index Measure (SSIM).
In our case, the MSE, defined as in following equation, was chosen. 
\begin{equation}
\mathcal{L}_{\text{MSE}} = \frac{1}{N} \sum_{i=1}^N \left( x_i - \hat{x}_i \right)^2
\label{MSE}
\end{equation}
where \(x_i\) is the \(i\)-th pixel of the input image, \(\hat{x}_i\) is the corresponding pixel in the reconstructed image, and \(N\) is the total number of pixels.
This type of loss is ideal for tasks requiring pixel-level accuracy.

\subsubsection{Alternative reconstruction models}
For completeness, we also tested an alternative reconstruction architecture based on the generator of Pix2Pix~\cite{isola2017image}, which employs a U-Net structure with skip connections and adversarial training.
Despite its strong generalization capabilities in image-to-image translation tasks, Pix2Pix yielded inferior performance in the reconstruction-based TTA framework.
In the LDCT denoising task, the $\mathrm{TTA}_{\mathrm{Grid}}$ configuration with Pix2Pix achieved lower results (SSIM: $.671 \pm .285$, MAE: $.086 \pm .074$, PSNR: $26.201 \pm 5.761$) compared to our autoencoder-based reconstruction models.
We hypothesize that Pix2Pix’s tendency to generalize across domains reduces its sensitivity to subtle distribution shifts, making it less suitable for TTA, where fine-grained domain discrepancies must be identified and corrected.
Based on these observations, we opted to use autoencoders as reconstruction models throughout our  experiments.

\subsection{Adaptors}
The adaptors are implemented as individual $1 \times 1$ convolutional layers.
Each feature-level adaptor consists of a $1 \times 1$ convolution that preserves the spatial resolution of the feature maps while altering their channel-wise representation.
The input-level adaptor, in contrast, operates directly on the input image using a sequence of convolutions that maintain its spatial dimensions.
In both cases, the use of $1 \times 1$ convolutions ensures that the spatial structure of the image remains unchanged, while enabling pixel-wise modulation of feature information.
Adaptors are placed at each level corresponding to a reconstruction model, with the exception of the final output layer, which remains unmodified.
Their purpose is to apply sample-specific transformations aimed at reducing domain shift, as estimated by the reconstruction error computed from the reconstruction models.

\newpage
\section{Sensitivity Analysis on Threshold Selection}
\label{Appendix:Threshold Selection}
To complement the results presented in the main manuscript (based on the $95^{th}$ percentile threshold), we report here the performance metrics obtained using alternative thresholds of $85^{th}$, $90^{th}$ and $98^{th}$ percentiles.
These thresholds were applied to determine the subset of test samples requiring TTA, denoted as $\bm{B} \subset \bm{A}$. 
It is worth noting that, due to rounding to four decimal places, the $85^{th}$ and $90^{th}$ percentiles resulted in the same numerical threshold on the BraTS dataset.

The results show that selecting $\tau$ equal to the $95^{th}$ percentile provides the best overall trade-off.  
On the one hand, lower values of $\tau$ yield a more liberal selection strategy, triggering adaptation for a larger portion of the test set.  
While this can enhance performance on highly shifted samples, it also increases the risk of unnecessary adaptation on ID data, potentially leading to performance degradation due to overfitting or instability.  
On the other hand, higher values of $\tau$ result in a conservative approach, limiting adaptation to only the most anomalous cases.  
This minimizes the risk of harming ID performance but may miss moderately shifted OOD samples that could benefit from adaptation.
Furthermore, since TTA involves non-negligible computational overhead, overly permissive thresholds, i.e., low $\tau$ values, may incur unnecessary cost by adapting a large number of samples that do not significantly benefit from it.  
Therefore, the $95^{th}$ percentile threshold emerges as a reasonable and robust compromise between effectiveness, selectivity, and computational efficiency.

For the $T_1$ to $T_2$ MRI translation task, the IXI dataset was used solely as an external test set for OOD evaluation.  
Therefore, we did not perform a sensitivity analysis on this dataset.  
Instead, we applied the threshold selected on the BraTS 2018 dataset ensuring that the choice is not biased by the test data distribution and mimics a realistic deployment setting.

\bb
\subsection{Denoising task on LDCT dataset}

 % thr = 85
\begin{table}[htpb]
\centering
\resizebox{.98\textwidth}{!}{%
\begin{tabular}{l c c c c c c c}
\toprule
\multicolumn{2}{c}{\textbf{Experiments}} 
& \multicolumn{2}{c}{\textbf{SSIM ↑}} 
& \multicolumn{2}{c}{\textbf{MAE ↓}} 
& \multicolumn{2}{c}{\textbf{PSNR ↑}} \\
\cmidrule(lr){1-2} \cmidrule(lr){3-4} \cmidrule(lr){5-6} \cmidrule(lr){7-8}
\textbf{} & \textbf{} 
& \textbf{$\bm A$} & $\bm{B \subset A}$  
& \textbf{$\bm A$} & $\bm{B \subset A}$  
& \textbf{$\bm A$} & $\bm{B \subset A}$  \\
\midrule
\multirow{3}{*}{Competitors} 
  & No TTA & .693 ± .199 & .623 ± .234$^\dagger$ & .063 ± .048 & .109 ± .082$^\dagger$ & 27.913 ± 5.793 & 20.396 ± 3.689$^\dagger$ \\
  & He et al.~\cite{he2021autoencoder} & .384 ± .167 & .459 ± .159$^\dagger$ & .388 ± .105 & .379 ± .112$^\dagger$& 15.723 ± 1.524 & 15.816 ± 1.440$^\dagger$ \\
  & He et al.~\textsuperscript{*} & .645 ± .211 & .643 ± .269$^\dagger$ & .089 ±.069 & .103 ± .070$^\dagger$  & 24.875 ± 4.277 & 24.280 ± 5.372$^\dagger$\\
\midrule
%\rowcolor{lightblue}
\multirow{6}{*}{Our Approach} 
  & $\mathrm{TTA}_{\mathrm{Grid}}$ &\cellcolor{lightgreen} .701 ± .205 &\cellcolor{lightgreen} .679 ± .278 & \cellcolor{lightgreen}.060 ± .042 &\cellcolor{lightgreen} .083 ± .070 & \cellcolor{lightgreen}28.808 ± 5.310 &\cellcolor{lightgreen} 26.839 ± 6.013 \\
  & $\mathrm{TTA}_{\mathrm{R10}}$*  & .699 ± .204 & .662 ± .271 & .061 ± .044 & .093 ± .075 & 28.610 ± 5.305 & 25.414 ± 5.330 \\
  & $\mathrm{TTA}_{\mathrm{R50}}$*  & .701 ± .205 & .680 ± .277 & .060 ± .042 & .083 ± .068 & 28.783 ± 5.293 & 26.664 ± 5.843 \\
  & $\mathrm{TTA}_{\mathrm{FS}}$  & .698 ± .205 & .661 ± .276 & .061 ± .046 & .094 ± .084 & 28.614 ± 5.392 & 25.445 ± 5.944 \\
  & $\mathrm{TTA}_{\mathrm{BE}}$ & .700 ± .204 & .671 ± .271 & .060 ± .042 & .087 ± .070 & 28.699 ± 5.304 & 26.047 ± 5.657 \\
& $\mathrm{TTA}_{\mathrm{BA}}$ & .696 ± .204 & .646 ± .267 & .050 ± .094 & .105 ± .094 & 28.449 ± 5.404 & 24.253 ± 5.239 \\
\bottomrule
\end{tabular}%
}
\caption[Results for denoising task on LDCT dataset using $85^{th}$ percentile as a threshold ]{Results for the denoising task on the LDCT dataset using the $85^{th}$ percentile as the threshold threshold for OOD detection. 
An asterisk (*) denotes that results were averaged over three runs to ensure reproducibility. 
Metrics are reported for the entire test set $\bm{A}$ and for the subset $\bm{B} \subset \bm{A}$. 
The best-performing configuration is highlighted in green.
The dagger symbol $^\dagger$ marks results from configurations that do not support OOD sample identification. 
Nonetheless, we report their performance on the subset $\bm{B} \subset \bm{A}$, as identified by our approach, to show how these configurations perform on detected OOD samples.}
\label{tab:ldct_thr85}
\end{table}
\FloatBarrier

 % thr = 90
 
\begin{table}[htpb]
\centering
\resizebox{.95\textwidth}{!}{%
\begin{tabular}{l c c c c c c c}
\toprule
\multicolumn{2}{c}{\textbf{Experiments}} 
& \multicolumn{2}{c}{\textbf{SSIM ↑}} 
& \multicolumn{2}{c}{\textbf{MAE ↓}} 
& \multicolumn{2}{c}{\textbf{PSNR ↑}} \\
\cmidrule(lr){1-2} \cmidrule(lr){3-4} \cmidrule(lr){5-6} \cmidrule(lr){7-8}
\textbf{} & \textbf{} 
& \textbf{$\bm A$} & $\bm{B \subset A}$  
& \textbf{$\bm A$} & $\bm{B \subset A}$  
& \textbf{$\bm A$} & $\bm{B \subset A}$  \\
\midrule
\multirow{3}{*}{Competitors} 
  & No TTA & .693 ± .199 & .667 ± .232$^\dagger$ & .063 ± .048 & .103 ± .069$^\dagger$ & 27.913 ± 5.793 & 19.690 ± 3.064$^\dagger$ \\
  & He et al.~\cite{he2021autoencoder} & .384 ± .167 & .502 ± .121$^\dagger$ & .388 ± .105 & .362 ± .091$^\dagger$& 15.723 ± 1.524 & 16.020 ± 1.359$^\dagger$ \\
  & He et al.~\textsuperscript{*} & .645 ± .211 & .709 ± .259$^\dagger$ & .089 ±.069 & .088 ± .072$^\dagger$  & 24.875 ± 4.277 & 25.177 ± 5.305$^\dagger$\\
\midrule
%\rowcolor{lightblue}
\multirow{6}{*}{Our Approach} 
  & $\mathrm{TTA}_{\mathrm{Grid}}$ &\cellcolor{lightgreen} .701 ± .204 &\cellcolor{lightgreen} .744 ± .271 &\cellcolor{lightgreen} .060 ± .044 &\cellcolor{lightgreen} .070 ± .057 &\cellcolor{lightgreen} 28.743 ± 5.365 &\cellcolor{lightgreen} 28.093 ± 5.966 \\
  & $\mathrm{TTA}_{\mathrm{R10}}$*  & .699 ± .203 & .732 ± .265 & .061 ± .044 & .076 ± .052 & 28.612 ± 5.335 & 26.770 ± 5.390 \\
  & $\mathrm{TTA}_{\mathrm{R50}}$*  & .701 ± .203 & .743 ± .270 & .060 ± .045 & .070 ± .053 & 28.723 ± 5.341 & 27.891 ± 5.714 \\
  & $\mathrm{TTA}_{\mathrm{FS}}$  & .699 ± .204 & .727 ± .267 & .061 ± .045 & .078 ± .063 & 28.584 ± 5.366 & 26.490 ± 5.597 \\
  & $\mathrm{TTA}_{\mathrm{BE}}$ & .700 ± .203 & .734 ± .263 & .060 ± .045 & .075 ± .052 & 28.633 ± 5.353 & 26.978 ± 5.629 \\
  & $\mathrm{TTA}_{\mathrm{BA}}$ & .696 ± .203 & .704 ± .261 & .062 ± .050 & .095 ± .090 & 28.437 ± 5.404 & 24.994 ± 5.308 \\
\bottomrule
\end{tabular}%
}
\caption[Results for denoising task on LDCT dataset using $90^{th}$ percentile as a threshold]{Results for the denoising task on the LDCT dataset using the $90^{th}$ percentile as threshold for OOD detection. 
An asterisk (*) denotes that results were averaged over three runs to ensure reproducibility. 
Metrics are reported for the entire test set $\bm{A}$ and for the subset $\bm{B} \subset \bm{A}$. 
The best-performing configuration is highlighted in green.
The dagger symbol $^\dagger$ marks results from configurations that do not support OOD sample identification. 
Nonetheless, we report their performance on the subset $\bm{B} \subset \bm{A}$, as identified by our approach, to show how these configurations perform on detected OOD samples.}
\label{tab:ldct_thr90}
\end{table}
\FloatBarrier

% thr= 98
\begin{table}[htpb]
\centering
\resizebox{.95\textwidth}{!}{%
\begin{tabular}{l c c c c c c c}
\toprule
\multicolumn{2}{c}{\textbf{Experiments}} 
& \multicolumn{2}{c}{\textbf{SSIM ↑}} 
& \multicolumn{2}{c}{\textbf{MAE ↓}} 
& \multicolumn{2}{c}{\textbf{PSNR ↑}} \\
\cmidrule(lr){1-2} \cmidrule(lr){3-4} \cmidrule(lr){5-6} \cmidrule(lr){7-8}
\textbf{} & \textbf{} 
& \textbf{$\bm A$} & $\bm{B \subset A}$  
& \textbf{$\bm A$} & $\bm{B \subset A}$  
& \textbf{$\bm A$} & $\bm{B \subset A}$  \\
\midrule
\multirow{3}{*}{Competitors} 
  & No TTA & .693 ± .199 & .706 ± .123$^\dagger$ & .063 ± .048 & .105 ± .038$^\dagger$ & 27.913 ± 5.793 & 17.660 ± 1.265$^\dagger$ \\
  & He et al.~\cite{he2021autoencoder} & .384 ± .167 & .563 ± .074$^\dagger$ & .388 ± .105 & .308 ± .044$^\dagger$& 15.723 ± 1.524 & 16.845 ± 1.151$^\dagger$ \\
  & He et al.~\textsuperscript{*} & .645 ± .211 & .840 ± .156$^\dagger$ & .089 ±.069 & .065 ± .043$^\dagger$  & 24.875 ± 4.277 & 27.395 ± 5.459$^\dagger$\\
\midrule
%\rowcolor{lightblue}
\multirow{6}{*}{Our Approach} 
  & $\mathrm{TTA}_{\mathrm{Grid}}$ &\cellcolor{lightgreen} .697 ± .202 & \cellcolor{lightgreen}.895 ± .158 &\cellcolor{lightgreen} .062 ± .047 &\cellcolor{lightgreen} .039 ± .030 &\cellcolor{lightgreen} 28.203 ± 5.663 &\cellcolor{lightgreen} 32.256 ± 4.151 \\
  & $\mathrm{TTA}_{\mathrm{R10}}$*  & .697 ± .201 & .881 ± .156 & .063 ± .048 & .047 ± .031 & 28.163 ± 5.641 & 30.222 ± 4.196 \\
  & $\mathrm{TTA}_{\mathrm{R50}}$*  & .697 ± .202 & .894 ± .158 & .062 ± .048 & .040 ± .032 & 28.195 ± 5.659 & 31.827 ± 4.301 \\
  & $\mathrm{TTA}_{\mathrm{FS}}$  & .696 ± .201 & .863 ± .162 & .063 ± .048 & .053 ± .038 & 28.138 ± 5.653 & 28.964 ± 5.266 \\
  & $\mathrm{TTA}_{\mathrm{BE}}$ & .696 ± .201 & .875 ± .158 & .063 ± .047 & .049 ± .035 & 28.164 ± 5.649 & 30.292 ± 4.669 \\
  & $\mathrm{TTA}_{\mathrm{BA}}$ & .696 ± .201 & .846 ± .156 & .06 ± .074 & .067 ± .041 & 28.112 ± 5.644 & 27.699 ± 4.751 \\
\bottomrule
\end{tabular}%
}
\caption[Results for denoising task on LDCT dataset using $98^{th}$ percentile as a threshold]{Results for denoising task on LDCT dataset using the $98^{th}$ percentile as threshold for OOD detection. 
An asterisk (*) denotes that results were averaged over three runs to ensure reproducibility. 
Metrics are reported for the entire test set $\bm{A}$ and for the subset $\bm{B} \subset \bm{A}$. 
The best results in each column are highlighted in green.
The dagger symbol $^\dagger$ marks results from configurations that do not support OOD sample identification. 
Nonetheless, we report their performance on the subset $\bm{B} \subset \bm{A}$, as identified by our approach, to show how these configurations perform on detected OOD samples.}
\end{table}
\FloatBarrier

\subsection{$T_1$-$T_2$ translation task on BraTS 2018}
% thr= 85
\begin{table}[!htb]
\centering
\resizebox{.95\textwidth}{!}{%
\begin{tabular}{l c c c c c c c}
\toprule
\multicolumn{2}{c}{\textbf{Experiments}} 
& \multicolumn{2}{c}{\textbf{SSIM ↑}} 
& \multicolumn{2}{c}{\textbf{MAE ↓}} 
& \multicolumn{2}{c}{\textbf{PSNR ↑}} \\
\cmidrule(lr){1-2} \cmidrule(lr){3-4} \cmidrule(lr){5-6} \cmidrule(lr){7-8}
\textbf{} & \textbf{} 
& \textbf{$\bm A$} & $\bm{B \subset A}$  
& \textbf{$\bm A$} & $\bm{B \subset A}$  
& \textbf{$\bm A$} & $\bm{B \subset A}$  \\
\midrule
%\rowcolor{lightgreen}
\multirow{3}{*}{Competitors} 
  & No TTA  &\cellcolor{lightgreen} .858 ± .040 & \cellcolor{lightgreen}.857 ± .025$^\dagger$ &\cellcolor{lightgreen} .037 ± .010 &\cellcolor{lightgreen} .040 ± .006$^\dagger$ &\cellcolor{lightgreen} 26.459 ± 1.864 &\cellcolor{lightgreen} 25.408 ± .716$^\dagger$ \\
  & He et al.~\cite{he2021autoencoder} & .330 ± .519 & .292± .042$^\dagger$ & .571 ± .060 & .060 ± .055$^\dagger$ & 7.23 ± .899 & 7.719 ± .803$^\dagger$\\
  & He et al.~\textsuperscript{*} & .803 ± .045 & .799± .033$^\dagger$  & .079 ± .028 & .073 ± .020$^\dagger$  & 20.498 ± 2.760 & 20.115 ± 2.106$^\dagger$  \\
\midrule
%\rowcolor{lightblue}
\multirow{6}{*}{Our Approach} 
  & $\mathrm{TTA}_{\mathrm{Grid}}$ & .853 ± .044 & .806 ± .034 & .039 ± .014 & .062 ± .018 & 26.034 ± 2.540 & 21.163 ± 2.258 \\
  & $\mathrm{TTA}_{\mathrm{R10}}$*  & .854 ± .043 & .810 ± .032 & .039 ± .013 & .058 ± .017 & 26.106 ± 2.409 & 21.886 ± 2.283 \\
  & $\mathrm{TTA}_{\mathrm{R50}}$*  & .853 ± .044 & .804 ± .035 & .040 ± .015 & .065 ± .020 & 26.032 ± 2.554 & 21.144 ± 2.370 \\
  & $\mathrm{TTA}_{\mathrm{FS}}$  & .853 ± .044 & .806 ± .040 & .039 ± .013 & .058 ± .017 & 26.084 ± 2.493 & 21.662 ± 2.719 \\
  & $\mathrm{TTA}_{\mathrm{BE}}$ & .853 ± .044 & .805 ± .035 & .040 ± .014 & .063 ± .018 & 26.031 ± 2.523 & 21.131 ± 1.977 \\
 & $\mathrm{TTA}_{\mathrm{BA}}$ & .853 ± .044 & .807 ± .035 & .040 ± .015 & .065 ± .020 & 26.042 ± 2.545 & 21.247 ± 2.486 \\
\bottomrule
\end{tabular}%
}
\caption[Results for $T_1-T_2$ translation task on BraTS 2018 dataset using $85^{th}$ percentile as a threshold]{Results for $T_1$-$T_2$ translation task on BraTS 2018 dataset using the $85^{th}$ percentile as threshold for OOD detection.
An asterisk (*) denotes that results were averaged over three runs to ensure reproducibility. 
Metrics are reported for the entire test set $\bm{A}$ and for the subset $\bm{B} \subset \bm{A}$. 
The best results in each column are highlighted in green.
The dagger symbol $^\dagger$ marks results from configurations that do not support OOD sample identification. 
Nonetheless, we report their performance on the subset $\bm{B} \subset \bm{A}$, as identified by our approach, to show how these configurations perform on detected OOD samples.
}
\end{table}
\FloatBarrier

% thr = 98
\begin{table}[!htb]
\centering
\resizebox{.95\textwidth}{!}{%
\begin{tabular}{l c c c c c c c}
\toprule
\multicolumn{2}{c}{\textbf{Experiments}} 
& \multicolumn{2}{c}{\textbf{SSIM ↑}} 
& \multicolumn{2}{c}{\textbf{MAE ↓}} 
& \multicolumn{2}{c}{\textbf{PSNR ↑}} \\
\cmidrule(lr){1-2} \cmidrule(lr){3-4} \cmidrule(lr){5-6} \cmidrule(lr){7-8}
\textbf{} & \textbf{} 
& \textbf{$\bm A$} & $\bm{B \subset A}$  
& \textbf{$\bm A$} & $\bm{B \subset A}$  
& \textbf{$\bm A$} & $\bm{B \subset A}$  \\
\midrule
%\rowcolor{lightgreen}
\multirow{3}{*}{Competitors} 
  & No TTA &\cellcolor{lightgreen} .858 ± .040 & \cellcolor{lightgreen} .862 ± .011$^\dagger$ &\cellcolor{lightgreen} .037 ± .010 &\cellcolor{lightgreen} .040 ± .004$^\dagger$ &\cellcolor{lightgreen} 26.459 ± 1.864 &\cellcolor{lightgreen} 24.887 ± .734$^\dagger$ \\
  & He et al.~\cite{he2021autoencoder} & .330 ± .519 & .273± .026 $^\dagger$ & .571 ± .060 & .061 ± .042$^\dagger$ & 7.23 ± .899 & 7.789 ± .924$^\dagger$\\
  & He et al.~\textsuperscript{*}  & .803 ± .045 & .788± .016$^\dagger$  & .079 ± .028 & .080 ± .012$^\dagger$  & 20.498 ± 2.760 & 18.623 ± 1.525$^\dagger$  \\
\midrule
%\rowcolor{lightblue}
\multirow{6}{*}{Our Approach} 
 % & $\mathrm{TTA}_{\mathrm{UP}}$ &\cellcolor{lightblue}  .858 ± .040 &\cellcolor{lightblue}  \textbf{.864 ± .012} &\cellcolor{lightblue} .037 ± .010 &\cellcolor{lightblue}  \textbf{.037 ± .004} & \cellcolor{lightblue} 26.474 ± 1.855 & \cellcolor{lightblue} \textbf{25.506 ± .848} \\ 
  & $\mathrm{TTA}_{\mathrm{Grid}}$ & .857 ± .041 & .791 ± .018 & .038 ± .012 & .074 ± .013 & 26.354 ± 2.103 & 19.366 ± 1.522 \\
  & $\mathrm{TTA}_{\mathrm{R10}}$*  & .857 ± .041 & .807 ± .016 & .038 ± .011 & .051 ± .006 & 26.417 ± 1.929 & 22.553 ± 1.032 \\
  & $\mathrm{TTA}_{\mathrm{R50}}$*  & .857 ± .041 & .790 ± .019 & .038 ± .012 & .076 ± .014 & 26.362 ± 2.084 & 19.762 ± 1.946 \\
  & $\mathrm{TTA}_{\mathrm{FS}}$  & .857 ± .041 & .793 ± .028 & .038 ± .011 & .066 ± .020 & 26.368 ± 2.098 & 20.080 ± 3.383 \\
  & $\mathrm{TTA}_{\mathrm{BE}}$ & .857 ± .041 & .789 ± .018 & .038 ± .012 & .077 ± .016 & 26.359 ± 2.090 & 19.608 ± 1.727 \\
   & $\mathrm{TTA}_{\mathrm{BA}}$ & .857 ± .041 & .802 ± .021 & .0438 ± .011 & .066 ± .016 & 26.390 ± 2.059 & 21.185 ± 2.067 \\
\bottomrule
\end{tabular}%
}
\caption[Results for $T_1-T_2$ translation task on BraTS 2018 dataset using $98^{th}$ percentile as a threshold]{Results for $T_1$-$T_2$ translation task on BraTS 2018 dataset using the $98^{th}$ percentile as threshold for OOD detection.
An asterisk (*) denotes that results were averaged over three runs to ensure reproducibility. 
Metrics are reported for the entire test set $\bm{A}$ and for the subset $\bm{B} \subset \bm{A}$. 
The best results in each column are highlighted in green.
The dagger symbol $^\dagger$ marks results from configurations that do not support OOD sample identification. 
Nonetheless, we report their performance on the subset $\bm{B} \subset \bm{A}$, as identified by our approach, to show how these configurations perform on detected OOD samples.
}
\end{table}
\FloatBarrier

\newpage
\section{Alternative search strategies}
To complement the exhaustive grid search used in our main implementation, we evaluated four additional strategies for selecting the optimal subset of reconstruction modules for each OOD test sample.
All these strategies operate within the same sample-aware TTA framework and aim to reduce the computational burden of adaptation while preserving its dynamic, sample-specific nature.

\paragraph{Random Search}
Algorithm~\ref{alg:random} implements a random sampling scheme in which only a fixed number $N_{\text{config}}$ of configurations are sampled from the full search space $\Omega$. 
This significantly reduces computational cost compared to exhaustive search. 
However, the resulting configuration may be suboptimal, as random sampling can fail to explore informative regions of the space, especially for small $N_{\text{config}}$.
In our experiments, we explore this strategy with $N_{\text{config}} = 10$ and $N_{\text{config}} = 50$.

\paragraph{Forward Selection}
Algorithm~\ref{alg:Fs} introduces a greedy strategy that starts from a minimal configuration containing only the fixed reconstruction modules at the input and output levels ($\mathcal{R}_x$ and $\mathcal{R}_y$). The set of selected intermediate reconstruction models, $\bm{\mathcal{R}}_{\text{sel}}$, is initially empty and is iteratively expanded by adding one candidate model at a time. At each step, the configuration yielding the lowest reconstruction error $\mathcal{\epsilon}_y = | \hat{\bm{y}}^a - \hat{\bm{y}}^a_r |$ is retained, and the process stops when no further improvement is observed.

\paragraph{Backward Elimination}
In contrast, Algorithm~\ref{alg:Bw} starts from a full configuration in which all intermediate reconstruction models are included, i.e., $\bm{\mathcal{R}_{\text{sel}}} = \bm{\mathcal{R}}$.
At each step, the algorithm evaluates reduced configurations obtained by removing one model at a time.
If a reduced configuration yields lower reconstruction error, it is adopted as the new candidate.
The process terminates once further removals degrade performance, and the best-performing configuration $\omega^\ast$ is retained.

\paragraph{Bayesian Optimization}
Finally, Algorithm~\ref{alg:bayes} implements a model-based strategy that builds a surrogate of the objective function to guide the search.
The first $n_{\text{start}}=5$ configurations are sampled randomly to initialize the history of evaluations. 
Subsequent $n_{\text{trials}} - n_{\text{start}}$, with $n_{\text{trials}}=20$ iterations are driven by a Tree-structured Parzen Estimator (TPE) sampler, a probabilistic model that balances exploration and exploitation \cite{bergstra2011algorithms, bergstra2013making, watanabe2023tree}. 
Like the previous methods, the goal is to identify the subset of reconstruction models that minimizes the reconstruction error $\mathcal{\epsilon}_y$.

\begin{algorithm}[h]
\small
\caption{Random Search}\label{alg:random}
\begin{algorithmic}[1]
\If {$\mathcal{\epsilon}_{\text{y}} > \tau $} 
\State $ \mathcal{\epsilon}^{\text{best}} \gets \infty$, $\omega^* \gets \text{None}$
\ForAll {$\omega \in \text{RandomSubset}(\Omega, \text{$N_{config}$})$} \Comment{Select $N_{config}$ random} \State \hfill \text{configurations from $\Omega$}
            \State $ \mathcal{\epsilon}_{\text{steps}}^{\text{best}} \gets \infty$ \Comment{Initialize best error for the current configuration}
        \For {$i = 1$ to M} \Comment{Iterate over adaptor update steps}
            \State $\bm{y}^a = \mathcal{T}^\omega(\bm{x})$ \Comment{Run the inference with the current configuration $\omega$}
            \State $\mathcal{\epsilon} \gets \mathcal{\epsilon}_y = \| \hat{\bm{y}}^a - \hat{\bm{y}}^a_r \|$ \Comment{Evaluate the output reconstruction error}
                     \If {$\mathcal{\epsilon} < \mathcal{\epsilon}_{\text{steps}}^{\text{best}}$}
                \State $ \mathcal{\epsilon}_{\text{steps}}^{\text{best}} \gets \mathcal{\epsilon}$ \Comment{Update best error value} 
                \State \hfill \text{for the current configuration $\omega$}
            \EndIf
        \EndFor
        \If {$ \mathcal{\epsilon}_{\text{steps}}^{\text{best}}  <  \mathcal{\epsilon}^{\text{best}}$} \Comment{If the current best error value is better than}
        \State \hfill \text{the overall best error value}
            \State $ \mathcal{\epsilon}^{\text{best}} \gets \mathcal{\epsilon}_{\text{steps}}^{\text{best}} $, $\omega^* \gets \omega$ \Comment{Update best configuration $\omega^\ast{}$}
        \EndIf
    \EndFor
\EndIf
\State \Return $\omega^*$ \Comment{Return best configuration $\omega^\ast{}$}
\end{algorithmic}
\end{algorithm}

\begin{algorithm}[ht]
\small
\caption{Forward Selection}\label{alg:Fs}
\begin{algorithmic}[1]
\If {$\mathcal{\epsilon}_{\text{y}} > \tau $} 
\State $ \mathcal{\epsilon}^{\text{best}} \gets \infty$, $\omega^* \gets \text{None}$
 \State $\bm{\mathcal{R}_{\text{sel}}} \gets \emptyset$ \Comment{Start the search without reconstruction models}
    \While {$(\bm{\mathcal{R}} - \bm{\mathcal{R}_{\text{sel}}}) \neq \emptyset$} \Comment{Continue the search while there are}
    \State \hfill \text{unselected reconstruction models}
    %\State \hfill reconstruction models
        \State $\omega = \emptyset$ \Comment{Initialize candidate configuration} 
        \State \hfill \text{with no reconstruction models}
        \For {$r \in (\bm{\mathcal{R}} - \bm{\mathcal{R}_{\text{sel}}})$} \Comment{Iterate over the unselected}
        \State \hfill \text{reconstruction models}
            \State $\omega \gets \omega     \cup \{r\}$ \Comment{ Update candidate configuration by adding}
            \State \hfill \text{reconstruction model $r$}
            \State $ \mathcal{\epsilon}_{\text{steps}}^{\text{best}} \gets \infty$ \Comment{Initialize best error for the current configuration}
            \For {$i = 1$ to M} \Comment{Iterate over adaptor update steps}
                \State $\bm{y}^a = \mathcal{T}^\omega(\bm{x})$ \Comment{Inference with the current configuration $\omega$}
                \State $\mathcal{\epsilon} \gets \mathcal{\epsilon}_y = \| \hat{\bm{y}}^a - \hat{\bm{y}}^a_r \|$ \Comment{Evaluate the output reconstruction error} 
            \If {$\mathcal{\epsilon} < \mathcal{\epsilon}_{\text{steps}}^{\text{best}}$}
                \State $ \mathcal{\epsilon}_{\text{steps}}^{\text{best}} \gets \mathcal{\epsilon}$ \Comment{Update best error value}
                \State \hfill \text{for the current configuration $\omega$}
                \EndIf
            \EndFor
        \If {$ \mathcal{\epsilon}_{\text{steps}}^{\text{best}}  <  \mathcal{\epsilon}^{\text{best}}$} \Comment{If the current best error value is better than}
        \State \hfill \text{the overall best error value}
            \State $ \mathcal{\epsilon}^{\text{best}} \gets \mathcal{\epsilon}_{\text{steps}}^{\text{best}} $, $\omega^* \gets \omega$ , $\bm{\mathcal{R}_{\text{sel}}}=\bm{\mathcal{R}_{\text{sel}}} \cup \{r\} $ \Comment{Update best} \State \hfill \text{configuration}
        \Else
            \State \textbf{break while} 
        \EndIf
        \EndFor
    \EndWhile
\EndIf
\State \Return $\omega^*$ \Comment{Return best configuration $\omega^\ast{}$}
\end{algorithmic}
\end{algorithm}

\FloatBarrier
\begin{algorithm}[ht]
\small
\caption{Backward Elimination}\label{alg:Bw}
\begin{algorithmic}[1]
\If {$\mathcal{\epsilon}_{\text{y}} > \tau $} 
\State $ \mathcal{\epsilon}^{\text{best}} \gets \infty$, $\omega^* \gets \text{None}$
\State  \( \bm{\mathcal{R}_{\text{sel}}} = \bm{\mathcal{R}} \) \Comment{Start the search with all reconstruction models}
    \While {$\bm{\mathcal{R}_{\text{sel}}} \neq \emptyset$} \Comment{Continue the search while there are}
    \State \hfill \text{reconstruction models in the selected set}
        \State $\omega = \bm{\mathcal{R}_{\text{sel}}}$ \Comment{Initialize candidate configuration}
        \State \hfill \text{with all reconstruction models}
        \For {$r \in \bm{\mathcal{R}_{\text{sel}}}$}  \Comment{Iterate over the selected reconstruction models}
            \State $\omega \gets \omega \setminus  \{r\}$ \Comment{Update the configuration by removing}
            \State \hfill \text{reconstruction model $r$}
            \State $ \mathcal{\epsilon}_{\text{steps}}^{\text{best}} \gets \infty$ \Comment{Initialize best error for the current configuration}
            \For {$i = 1$ to M} \Comment{Iterate over adaptor update steps}
                \State $\bm{y}^a = \mathcal{T}^\omega(\bm{x})$ \Comment{Inference with the current configuration $\omega$}
                \State $\mathcal{\epsilon} \gets  \mathcal{\epsilon}_y = \| \hat{\bm{y}}^a - \hat{\bm{y}}^a_r \|$ \Comment{Evaluate the output reconstruction error} 
            \If {$\mathcal{\epsilon} < \mathcal{\epsilon}_{\text{steps}}^{\text{best}}$}
                \State $ \mathcal{\epsilon}_{\text{steps}}^{\text{best}} \gets \mathcal{\epsilon}$ \Comment{Update best error value}
                \State \hfill \text{for the current configuration $\omega$}
                \EndIf
            \EndFor
         \If {$ \mathcal{\epsilon}_{\text{steps}}^{\text{best}}  <  \mathcal{\epsilon}^{\text{best}}$} \Comment{If the current best error value is better}
         \State \hfill \text{than the overall best error value}
            \State $ \mathcal{\epsilon}^{\text{best}} \gets \mathcal{\epsilon}_{\text{steps}}^{\text{best}} $, $\omega^* \gets \omega$, $\bm{\mathcal{R}_{\text{sel}}}= \bm{\mathcal{R}_{\text{sel}}} \setminus \{r\}$ \Comment{Update best}
            \State \hfill \text{configuration}
        \Else
            \State \textbf{break while} 
        \EndIf
        \EndFor
    \EndWhile
\EndIf
\State \Return $\omega^*$ \Comment{Return best configuration $\omega^\ast{}$}
\end{algorithmic}
\end{algorithm}
\FloatBarrier

\begin{algorithm}[ht]
\small
\caption{Bayesian Search}\label{alg:bayes}
\begin{algorithmic}[1]
\If {$\mathcal{\epsilon}_{\text{y}} > \tau $} 
  \State $ \mathcal{\epsilon}^{\text{best}} \gets \infty$, $\omega^* \gets \text{None}$
\For{$t = 1$ to $n_{\text{trials}}$} \Comment{Iterate over Bayesian search trials}
    \If{$t \leq n_{\text{start}}$}
      \State $\omega \gets \texttt{RandomSample()}$ \Comment{Randomly sample a configuration}
      \State \hfill \text{(initialization phase)}
    \Else
      \State $\omega \gets \texttt{Sampler}(\{ (\omega_i, \mathcal{\epsilon}_{\text{y},i}) \}_{i=1}^{t-1})$ \Comment{Select configuration using surrogate}
      \State \hfill \text{model based on past evaluations}
    \EndIf
    \State $ \mathcal{\epsilon}_{\text{steps}}^{\text{best}} \gets \infty$ \Comment{Initialize best error for the current configuration}

    \For{$i = 1$ to $M$} \Comment{Iterate over adaptor update steps}
      \State $\bm{y}^a = \mathcal{T}^\omega(\bm{x})$ \Comment{Inference with the current configuration $\omega$}
      \State $\mathcal{\epsilon} \gets \mathcal{\epsilon}_y = \| \hat{\bm{y}}^a - \hat{\bm{y}}^a_r \|$ \Comment{Evaluate the output reconstruction error} 
            \If {$\mathcal{\epsilon} < \mathcal{\epsilon}_{\text{steps}}^{\text{best}}$}
                \State $ \mathcal{\epsilon}_{\text{steps}}^{\text{best}} \gets \mathcal{\epsilon}$ \Comment{Update best error value}
                 \State \hfill \text{for the current configuration $\omega$}
            \EndIf
        \EndFor
        \If {$ \mathcal{\epsilon}_{\text{steps}}^{\text{best}}  <  \mathcal{\epsilon}^{\text{best}}$} \Comment{If the current best error value is better} 
         \State \hfill \text{than the overall best error value}
            \State $ \mathcal{\epsilon}^{\text{best}} \gets \mathcal{\epsilon}_{\text{steps}}^{\text{best}} $, $\omega^* \gets \omega$ \Comment{Update best configuration $\omega^\ast{}$}
        \EndIf
    \EndFor
\EndIf
\State \Return $\omega^*$ \Comment{Return best configuration $\omega^\ast{}$}
\end{algorithmic}
\end{algorithm}

\newpage

\section{Wilcoxon Test}
\label{Appendix: statistical Analysis}

To assess the statistical significance of performance differences between methods, we employ the Wilcoxon signed-rank test~\cite{wilcoxon1992individual}, a non-parametric test suitable for paired, non-normally distributed data.
It is particularly appropriate for our setting, as it compares the distributions of metric values (e.g., SSIM, MAE, PSNR) across all test samples without assuming Gaussianity, and it is robust to outliers.
We perform pairwise comparisons between all evaluated methods, independently for each metric, and apply a Bonferroni correction to account for the increased risk of Type I error due to multiple comparisons \cite{bonferroni1936teoria}.
We use a corrected significance threshold $\alpha_{\text{corr}} =\frac{0.05}{m}$ where $m$ is the number of pairwise comparisons.
Results are summarized in compact tables, where each cell corresponds to a pairwise comparison (row vs. column), and is subdivided into three subcells showing the $p$-values for SSIM, MAE, and PSNR.
Green subcells indicate statistically significant differences ($p < \alpha_{\text{corr}}$), while red subcells denote non-significant differences ($p \geq \alpha_{\text{corr}}$).
This analysis supports a robust comparison of all TTA configurations and highlights whether observed improvements are consistent and statistically reliable.

Wilcoxon test results are reported in Table~\ref{tab: wilcoxon_denoising}, and Table~\ref{tab: wilcoxon_IXI}, which correspond to the Mayo Clinic LDCT-and-Projection dataset~\cite{moen2021low}, and IXI dataset~\cite{IXI}, respectively.
No statistical test was performed on the BraTS 2018 dataset~\cite{menze2014multimodal}, as TTA consistently degrades performance in this predominantly ID scenario. 
Consequently, we did not explore additional search strategies in this setting.

\begin{table}[htpb]
\centering
\begin{tikzpicture}
    \node at (0,0) {\includegraphics[width=0.9\linewidth]{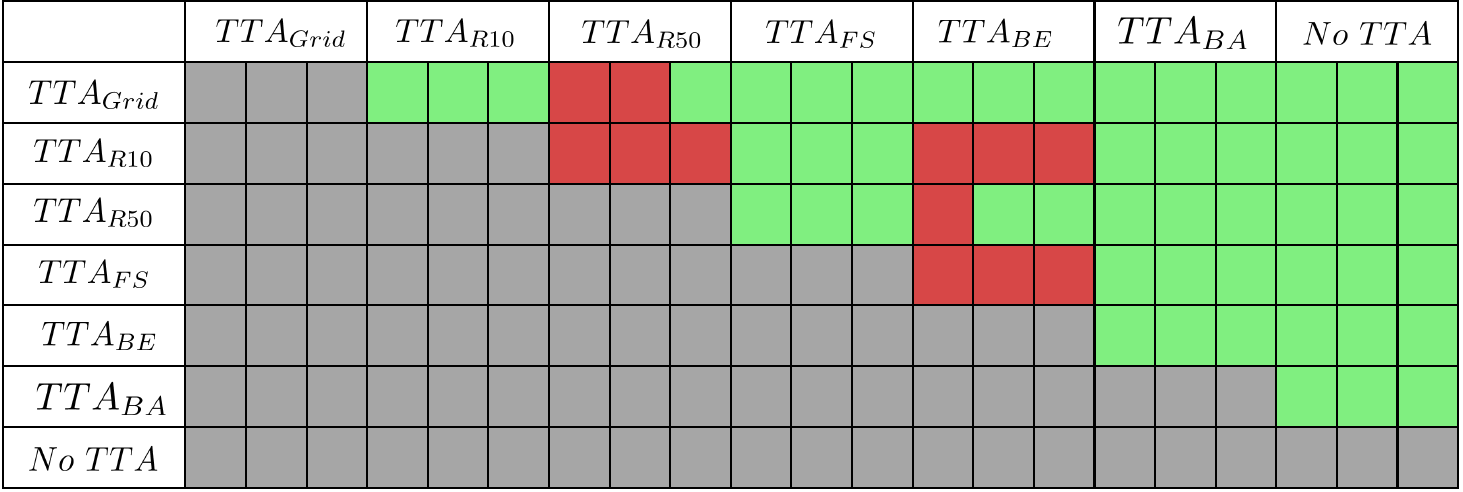}};
\end{tikzpicture}
 \caption[Wilcoxon test with Bonferroni correction for LDCT denoising task]{Wilcoxon test with Bonferroni correction for denoising task. This table summarizes the pairwise comparisons between different TTA configurations using the Wilcoxon signed-rank test, with Bonferroni correction applied to account for multiple comparisons. 
Each cell corresponds to a comparison between two methods (row vs. column) and is split into three subcells, reporting the statistical significance ($p$-value) for SSIM, MAE, and PSNR, respectively. Green subcells indicate statistically significant differences ($p < \alpha_{\text{corr}}$), while red ones indicate non-significant differences ($p \geq \alpha_{\text{corr}}$).}
    \label{tab: wilcoxon_denoising}
\end{table}
\FloatBarrier

\begin{table}[t]
\centering
\begin{tikzpicture}
    \node at (0,0) {\includegraphics[width=0.9\linewidth]{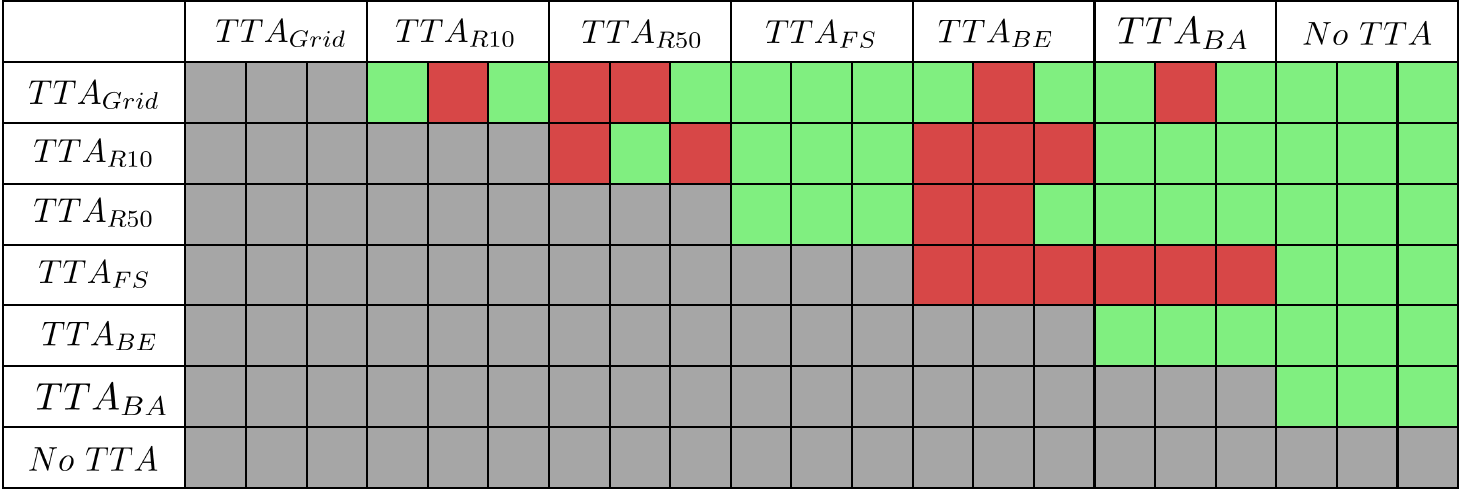}};
\end{tikzpicture}
 \caption[Wilcoxon test with Bonferroni correction for $T_1$-$T_2$ MRI translation on IXI dataset]{Wilcoxon test with Bonferroni correction for $T_1$-$T_2$ MRI translation on IXI dataset. This table summarizes the pairwise comparisons between different TTA configurations using the Wilcoxon signed-rank test, with Bonferroni correction applied to account for multiple comparisons. 
Each cell corresponds to a comparison between two methods (row vs. column) and is split into three subcells, reporting the statistical significance ($p$-value) for SSIM, MAE, and PSNR, respectively. Green subcells indicate statistically significant differences ($p < \alpha_{\text{corr}}$), while red ones indicate non-significant differences ($p \geq \alpha_{\text{corr}}$).}
   \label{tab: wilcoxon_IXI}
\end{table}
\FloatBarrier

\end{document}